\documentclass[12pt]{article}
\usepackage{amsmath}
\usepackage{graphicx,psfrag,epsf}
\usepackage{enumerate}
\usepackage{natbib}
\usepackage{url} 

\usepackage{enumerate}
\usepackage{natbib}
\usepackage{xcolor}
\usepackage{subcaption}
\usepackage{caption}
\usepackage{amsthm}
\usepackage{eufrak}
\usepackage{amssymb}
\usepackage{url}
\usepackage{bm}

\usepackage{amsmath}
\usepackage{graphicx,psfrag,epsf}
\usepackage{enumerate}
\usepackage{natbib}
\usepackage{xcolor}
\usepackage{subcaption}
\usepackage{amsthm}
\usepackage{eufrak}
\usepackage{amssymb}
\usepackage{rotating}

\usepackage{mathrsfs}
\usepackage[ruled,vlined]{algorithm2e}
\SetKwComment{Comment}{//}

\newtheorem*{proof*}{Proof}
\newtheorem{lemma}{Lemma}
\newtheorem{theorem}{Theorem}

\newcommand{\blind}{1}

\newcommand{\diag}[1]{\hbox{diag}(#1)}

\begin{document}

\def\spacingset#1{\renewcommand{\baselinestretch}%
{#1}\small\normalsize} \spacingset{1}


\if1\blind
{
  \title{\bf Bridging the Usability Gap: Theoretical and Methodological Advances for Spectral Learning of Hidden Markov Models}
  \author{Xiaoyuan Ma\\
    Department of Statistics, University of Virginia\\
    and \\
    Jordan Rodu \\
    Department of Statistics, University of Virginia}
  \maketitle
} \fi

\if0\blind
{
  \bigskip
  \bigskip
  \bigskip
  \begin{center}
    {\LARGE\bf Bridging the Usability Gap: Theoretical and Methodological Advances for Spectral Learning of Hidden Markov Models}
\end{center}
  \medskip
} \fi

\bigskip
\begin{abstract}
The Baum-Welch (B-W) algorithm is the most widely accepted method for inferring hidden Markov models (HMM). However, it is prone to getting stuck in local optima, and can be too slow for many real-time applications.  Spectral learning of HMMs (SHMM), based on the method of moments (MOM) has been proposed in the literature to overcome these obstacles. Despite its promise, SHMM has lacked a comprehensive asymptotic theory. Additionally, its long-run performance can degrade due to unchecked error propagation.  In this paper, we (1) provide an asymptotic distribution for the approximate error of the likelihood estimated by SHMM, (2) propose a novel algorithm called projected SHMM (PSHMM) that mitigates the problem of error propagation, and (3) describe online learning variants of both SHMM and PSHMM that accommodate potential nonstationarity. We compare the performance of SHMM with PSHMM and estimation through the B-W algorithm on both simulated data and data from real world applications, and find that PSHMM not only retains the computational advantages of SHMM, but also provides more robust estimation and forecasting. 
\end{abstract}

\noindent%
\vfill

\newpage
\spacingset{1.45} 

\section{Introduction} \label{sec:intro}
The hidden Markov model (HMM), introduced by \citet{baum1966statistical}, has found widespread applications across diverse fields. These include finance \citep{Hassan2005-ss, Mamon2014-bh}, natural language processing \citep{Zhou2001-sb, Stratos2016-wv}, and medicine \citep{Shirley2010-bf, Scott2005-ar}.  An HMM is a stochastic probabilistic model for sequential or time series data that assumes that the underlying dynamics of the data are governed by a Markov chain. \citep{knoll2016using}. 

A widely adopted method for inferring HMM parameters is the Baum-Welch algorithm \citep{baum1970maximization}. This algorithm is a special case of the expectation maximization (EM) algorithm \citep{Dempster1977-kw}, grounded in maximum likelihood estimation (MLE).  However, the E-M algorithm can require a large number of iterations until the parameter estimates converge--which has a large computational cost, especially for large-scale time series data--and can easily get trapped in local optima. To address these issues, particularly for large, high-dimensional time series, \citet{hsu2012spectral} proposed a spectral learning algorithm for HMMs (SHMM). This algorithm, based on the method of moments (MOM), offers attractive theoretical properties.  However, the asymptotic error distribution of the algorithm was not well characterized.  Later, \cite{rodu2014spectral} improved and extended the spectral estimation algorithm to HMMs with high-dimensional, continuously distributed output, but again did not address the asymptotic error distribution.  In this manuscript, we provide a theoretical discussion of the asymptotic error behavior of SHMM algorithms.

In addition to investigating the asymptotic error distribution, we provide a novel improvement to the SHMM family of algorithms.  Our improvement is motivated from an extensive simulation study of the methods proposed in \cite{hsu2012spectral} and \cite{rodu2014spectral}.  We found that spectral estimation does not provide stable results under the low signal-noise ratio setting. We propose a new spectral estimation method, the projected SHMM (PSHMM), that leverages a novel regularization technique that we call `projection-onto-simplex' regularization.  The PSHMM largely retains the computational advantages of SHMM methods without sacrificing accuracy. 

Finally, we show how to adapt spectral estimation (including all SHMM and PSHMM approaches) to allow for online learning.  We examine two approaches -- the first speeds up computational time required for learning a model in large data settings, and the second incorporates ``forgetfulness,'' which allows for adapting to changing dynamics of the data.  This speed and flexibility is crucial, for instance, in high-frequency trading, and we show the effectiveness of the PSHMM on real data in this setting.

The structure of this paper is as follows: In the rest of this section, we will introduce existing models. In Section \ref{sec:error bound}, we provide theorems for the asymptotic properties of SHMMs.  Section \ref{sec:regularized SHMM} introduces our new method, PSHMM.  In Section \ref{sec: online learning}, we extend the spectral estimation to online learning for both SHMM and PSHMM.  Then, Section \ref{sec:simulations} shows results from simulation studies and Section \ref{sec: application} shows the application on high-frequency trading. We provide extensive discussion in Section \ref{sec:discussion}.

\subsection{The hidden Markov model}
\begin{figure}[!hbt]
    \centering
    \includegraphics[width=5in]{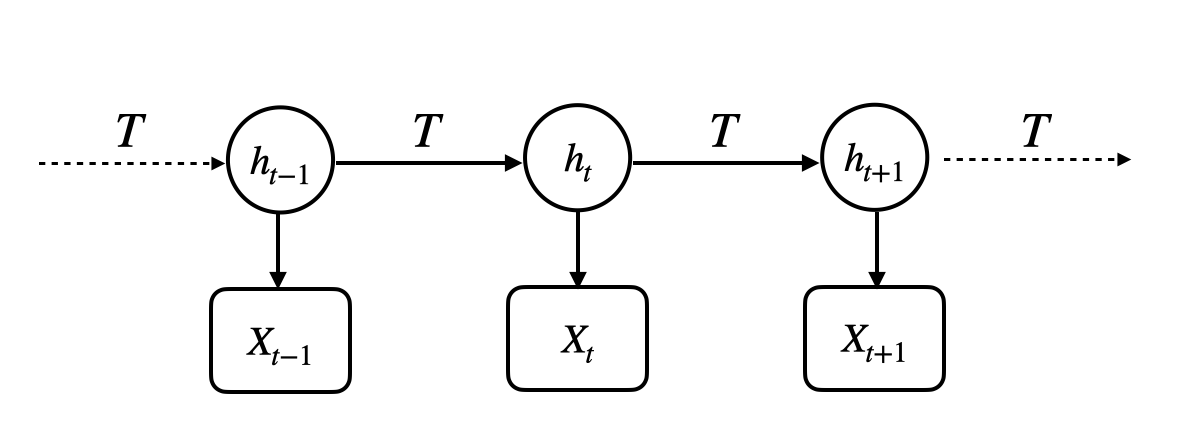}
    \caption{Model structure of standard HMM. $\{ h_t \} $ is a latent Markov chain that evolves according to transition matrix $\mathbf{T}$. For each time stamp $t$, the observed $X_t$ is generated according to the emission distribution associated with $h_t$.  } 
    \label{fig:standard HMM model structure}
\end{figure}

The standard HMM \citep{baum1966statistical} is defined by a set of $S$ hidden categorical states $1, 2, \cdots, S$ that evolve according to a Markov chain.  Let $h_t$ denote the hidden state at time $t$. The Markov chain is defined by two key components:
\begin{enumerate}
    \item an initial probability $\pi_0 = [\pi_0^{(1)}, \cdots, \pi_0^{(S)}]$ where $h_1 \sim Multinomial (\pi_0^{(1)}, \cdots, \pi_0^{(S)}) $, and
    \item a transition matrix $\mathbf{T} = [\mathbf{T}_{ij}]_{i=1,\cdots, S}^{j=1,\cdots, S}$ where $\mathbf{T}_{ij} = \mathrm{P} (h_{t+1} = j | h_{t} = i )$ for $\forall t$.
\end{enumerate}  
At each time point $t$, an observation $X_t$ is emitted from a distribution conditional on the value of the current hidden state, $X_t | h_t = s \sim \mathcal{F}_s$ where $\mathcal{F}_s$ is the emission distribution when the hidden state $h_t = s$.  Figure \ref{fig:standard HMM model structure} is a graphical representation of the standard HMM. When the emission distribution $\mathcal{F}_s$ is Gaussian for all states, the model is often called a Gaussian HMM (GHMM). 

\subsection{Spectral learning of HMM} \label{sec: SHMM_review}
\begin{figure}[!hbt]
    \centering
    \includegraphics[width=5in]{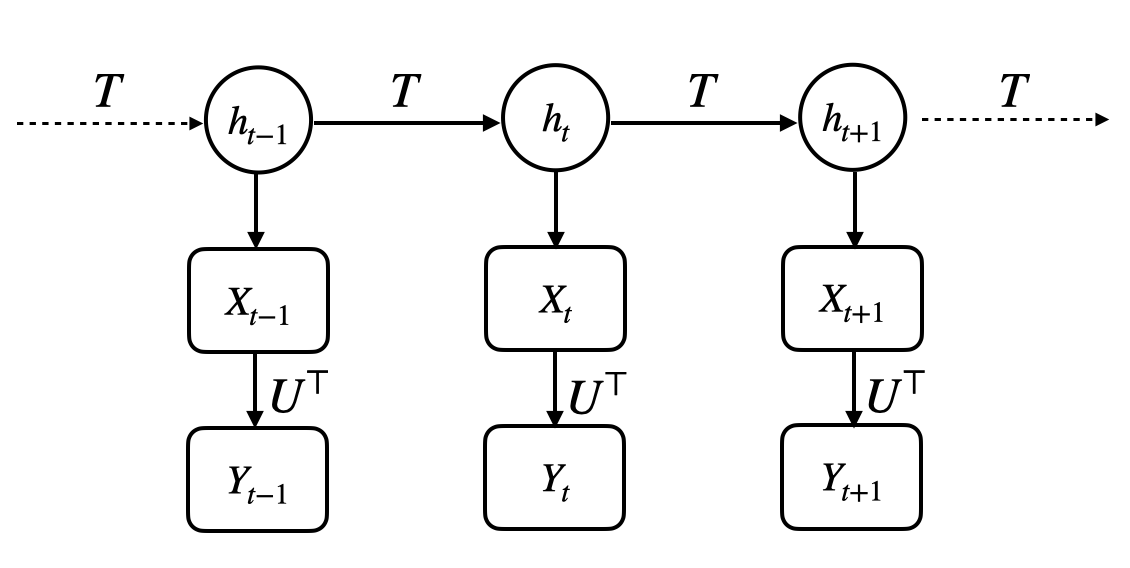}
    \caption{Spectral estimation model by \cite{rodu2014spectral}. In addition to the latent state series $\{ h_t \}_t$ and observed series $\{ X_t \}_t$, \citet{rodu2014spectral} introduced a reduced-dimensional series $ \{ Y_t = U^\top X_t \} $ which is a projection of $X_t$ on a lower-dimensional subspace whose dimensionality is equal to the number of hidden states. Spectral estimation proceeds based on $\{ Y_t \}_t$. } 
    \label{fig:SE_HMM_rodu} 
\end{figure}


Figure \ref{fig:SE_HMM_rodu} illustrates the model proposed by \cite{rodu2014spectral}. As in the standard HMM, $h_t$ denotes the hidden state at time $t$, and $X_t$ the emitted observation. To estimate the model, observations are projected onto a lower-dimensional space of dimensionality $d$.  We denote these lower-dimensional projected observations as $y_t$, where $y_t=U^\top x_t$. We address the selection of dimension $d$ and the method for projecting observations in Section \ref{subsec: pshmm hyperparameter choice}. 

The spectral estimation framework builds upon the observable operator model \citep{Jaeger2000-vj}.  In this model, the likelihood of a sequence of observations is expressed as:
\begin{eqnarray*}
    Pr(x_{1:t}) & = & 1^\top A(x_t)A(x_{t-1})\cdots A(x_1)\pi_0
\end{eqnarray*}
with $A(x_j) = T\diag{\lambda(x_j)}$ and $\lambda(x_j)$ is a vector with elements $[\lambda(x_j)]_i = \Pr(x_j | H_j = h_i)$.  Note that this formulation of $A(x)$ still requires inferring $T$ and $\lambda(x_j)$.  To address this, the spectral estimation framework employs a similarity transformation of $A(x)$.  This transformed version can be learned directly from the data using the MOM. 

Under the spectral estimation framework of \citet{hsu2012spectral}, the likelihood can be written as
\begin{align*}
P(x_{1:t}) &= 1^\top ~~ A(x_t) ~~\cdots ~~ A(x_1) ~~\pi_0\\
&= \underbrace{1^\top S^{-1}}_{b_\infty^\top} ~~ \underbrace{S A(x_t) S^{-1}}_{B(x_t)} ~~ S ~~\cdots ~~ S^{-1} ~~ S A(x_1) S^{-1} ~~  \underbrace{S \pi}_{b_1}\\
&\equiv b_\infty^\top ~~ B(x_t) ~~ \cdots ~~  B(x_1) ~~ b_1
\end{align*}
where $S=U^\top M$ with $M = [M_1 , \cdots, M_S]$ and column vector $ M_i = \mathrm{E} (X | H=i)$. Letting $y_t=U^\top x_t$, \citet{rodu2014spectral} shows that, the likelihood can be expressed as
\begin{eqnarray}
    Pr(x_{1:t}) & = & c_{\infty}^\top C(y_t)C(y_{t-1})\cdots C(y_1)c_1 ,
    \label{eq:rodu_lik}
\end{eqnarray}
 where
\begin{eqnarray*}
    & c_1 = \mu, c_{\infty}^\top = \mu^\top \Sigma^{-1}, C(y) = K(y)\Sigma^{-1},\\
    & \mu = E(y_1) = U^\top M \pi,\\
    & \Sigma = E(y_2 y_1^\top)=U^\top M T \diag{\pi} M^\top U,\\
    & K (a) = \mathrm{E}(y_3 y_1^\top y_2^\top) a = U^\top M T \diag{M^\top Ua} T \diag{\pi} (M^\top U) , \\
    & M = [M_1 , \cdots, M_S] \text{ where } M_i = \mathrm{E} (X | H=i) .
\end{eqnarray*}

These quantities can be empirically estimated as:
\begin{eqnarray}
    \widehat Pr(x_{1:t}) & = & \hat c_{\infty}^\top \hat C(y_t)\hat C(y_{t-1}) \cdots \hat C(y_1)\hat c_1 , 
\end{eqnarray}
where
\begin{eqnarray}
    & \hat  c_1 = \hat  \mu, \hat  c_{\infty}^\top = \hat  \mu^\top \hat  \Sigma^{-1}, \hat C(y) = \hat K(y)\hat \Sigma^{-1}, \nonumber\\
    & \hat \mu = \frac{1}{N} \sum_{i=1}^N Y_{i},\nonumber\\
    & \hat \Sigma = \frac{1}{N} \sum_{i=1}^{N-1} Y_{i+1} Y_{i}^\top,\nonumber\\
    & \hat K (y) = \frac{1}{N} \sum_{i=1}^{N-2} Y_{i+2}Y_{i}^\top \cdot Y_{i+1}^\top y. \label{eq: i.i.d triple estimation}
\end{eqnarray}

Prediction of $y_t$ is computed recursively by: 
\begin{eqnarray}
    \hat{y}_t & = & \frac{\hat{C} (y_{t-1}) \hat{y}_{t-1} }{\hat{c}^\top_{\infty} \hat{C} (y_{t-1}) \hat{y}_{t-1} } ;
    \label{eq:recursive_prediction_rodu}
\end{eqnarray}
Observation $x_t$ can be recovered as:
\begin{eqnarray}
    \hat{x_t} |x_1,x_2,\cdots,x_{t-1} = U \hat{y_t}.
\end{eqnarray}

In the above exposition of spectral likelihood estimation, moment estimation, and recursive forecasting we assume a discrete output HMM. For a continuous output HMM, the spectral estimation of likelihood is slightly different.  We need some kernel function $G(x)$ to calculate $K$, so $K (a) = U^\top M T \diag{M^\top U G(a)} T \diag{\pi} (M^\top U) $.

$G(a)$ provides a link between the observations and their conditional probabilities in the continuous output scenario.  Let $\tilde H_t$ be the probability vector associated with being in a particular state at time $t$, commonly called a \emph{belief state}.  Then $E[Y_2 | \tilde H_2] = U^\top M\tilde H_2$.  Therefore,
\begin{align*}
E[Y_2 | \tilde H_1] &= U^\top MT\tilde H_1.
\end{align*}   
Bayes formula gives
\begin{align*}
\Pr(H_1 | X_1) &= \frac{\Pr(X_1|H_1) \Pr(H_1)}{\Pr(X_1)}.
\end{align*}
Letting $[\lambda(x)]_i = \Pr(x_t | h_t = i)$, this implies that
\begin{align*}
E[H_1 | X_1=x] &= \frac{\diag{\pi}\lambda(x)}{\pi^\top \lambda(x)},
\end{align*}
Then we get
\begin{align*}
E[Y_2 | X_1=x] &= \frac{U^\top M T \diag{\pi}\lambda(x)}{\pi^\top \lambda(x)}
\end{align*}
Finally
\begin{align*}
\Sigma^{-1} E[Y_2 | X_1=x] &= \frac{(M^\top U)^{-1} \lambda(x)}{\pi^\top \lambda(x)}\\
&\equiv G(x)
\end{align*}
Using this definition of $G(x)$ we have 
\begin{align*}
K (x) &= U^\top M T \diag{M^\top U G(x)} T \diag{\pi} (M^\top U) \\
&=U^\top M T \mathrm{diag}\left(M^\top U \frac{(M^\top U)^{-1} \lambda(x)}{\pi^\top \lambda(x)}\right) T \diag{\pi} (M^\top U)\\
&= \frac{1}{{\pi^\top \lambda(x)}}U^\top M T \diag{\lambda(x)} T \diag{\pi} (M^\top U)
\end{align*}
Note from above that this is exactly what we want, up to a scaling constant that depends on $x$ \citep[for more on $G(\cdot)$ see][]{rodu2014spectral}. In this paper, we will use a linear kernel $G(a) = a$ for simplicity.  The moment estimation and recursive forecasting for the continuous case are identical to the discrete case. See \cite{rodu2013using} for detailed derivations and mathematical proofs of these results. 

In the following, the theoretical properties discussed in Section \ref{sec:error bound} are based on the likelihood (Eq \ref{eq:rodu_lik} - Eq \ref{eq: i.i.d triple estimation}), and improvements to prediction in Section \ref{sec:regularized SHMM} are based on Eq \ref{eq:recursive_prediction_rodu}. Note that deriving the prediction function from the likelihood is straightforward.  See \cite{rodu2014spectral} and \cite{hsu2012spectral} a detailed derivation.

\section{Theoretical Properties of SHMM}\label{sec:error bound}

\subsection{Assumptions for SHMMs}
The SHMM framework, (illustrated in Figure \ref{fig:SE_HMM_rodu}) relies on three primary assumptions: 
\begin{itemize}
    \item Markovian hidden states: The sequence of underlying hidden states $\{h_t\}$ follows a Markov process.
    \item Conditional independence: The observations $\{x_{1:T}\}$ are mutually independent, given the hidden states. 
    \item Invertibility condition: The matrix $\Sigma = E(y_2 y_1^\top)$ is invertible.  
\end{itemize}

The first two assumptions are standard assumptions for the HMM. The third assumption places constraints on the separability of the conditional distributions given the hidden states.  It also requires that the underlying transition matrix for the HMM is full rank.  While the full-rank condition is crucial for the standard SHMM we address in this paper, it is worth noting that some scenarios may involve non-full rank HMMs. For such cases, relaxations of this assumption have been explored in the literature \citep[e.g.][]{Siddiqi2009-py}.

\subsection{Asymptotic framework for SHMMs}
SHMM relies on the MOM to estimate the likelihood, offering a computationally efficient approach.  However, the asymptotic behavior of the SHMM has not been thoroughly explored in previous literature.  Prior work by \citet{hsu2012spectral} and \citet{rodu2013using} established conditions for convergence of the spectral estimator to the true likelihood almost surely, for discrete and continuous HMM respectively.  Specifically, they showed:
\begin{eqnarray}
    \widehat Pr(x_{1:T}) & = & \hat c_{\infty}^\top \hat C(y_t)\hat C(y_{t-1}) \cdots \hat C(y_1)\hat c_1 \xrightarrow[]{a.s.} Pr(x_{1:T}). 
    \label{eq: rodu_proof}
\end{eqnarray}
Here, $c_1, C(y_t)$ and $c_{\infty}$ are estimated using independent and identically distributed (i.i.d) triples $\{Y_i, Y_{i+1} , Y_{i+2}\}$, as defined in Eq \ref{eq: i.i.d triple estimation}. It is important to note that the use of i.i.d. triples is primarily a theoretical tool, used by both \cite{hsu2012spectral} and \cite{rodu2013using} to facilitate study of the theoretical properties of the SHMM.  In practice, quantities should be estimated using the full sequence of observations, which introduces dependencies between the triplets (see section \ref{sec:regularized SHMM}).  The relationship between the effective sample size for estimation and the actual sample size is governed by the mixing rate of the underlying Markov chain.

In this paper, we extend these results by investigating the asymptotic distribution of the estimation error, $\hat Pr(x_{1:T}) - Pr(x_{1:T})$. Theorem \ref{theorem: CLT} presents a CLT type bound for this approximation error.  We first identify the sources of error for the SHMM in Lemma \ref{lemma:likelihood_decomposition}, which makes use of what we refer to as the `$\Delta$' terms, defined through the following equations: $\widehat{\mu}  =  \mu + \widehat{\Delta \mu}$, $ \widehat{\Sigma} = \Sigma + \widehat{\Delta \Sigma}$, $ \widehat{K} = K +  \widehat{\Delta K}$. 

\begin{lemma}
    \begin{eqnarray}
     \widehat Pr(x_{1:T}) \nonumber & = &
     Pr(x_{1:T}) + (v+ \tilde{v})^\top \widehat{\Delta \mu} + \sum_{t=1}^T a_t^\top \widehat{\Delta K}(y_t) \tilde{a_t} - \sum_{t=0}^T b_t^\top \widehat{\Delta \Sigma} \tilde{b_t} + \mathcal{O}_p (N^{-1}) \label{SimplifiedExpansion},
\end{eqnarray}
where 
\begin{eqnarray}
     v =  \left( \mu^\top \Sigma^{-1} K(y_T) \cdots K(y_1) \Sigma^{-1} \right)^\top ;  & \quad
     \tilde{v} = \Sigma^{-1} K(y_T) \cdots K(y_1) \Sigma^{-1} \mu ; \nonumber\\ 
    a_t  = \left( \mu^\top \Sigma^{-1} K(y_T) \Sigma^{-1} \cdots K(y_{t+1})\Sigma^{-1} \right)^\top ; & \quad
    \tilde{a_t}  = \Sigma^{-1}K(y_{t-1}) \cdots K(y_1)\Sigma^{-1} \mu ; \nonumber \\ 
     b_t = \left(  \mu^\top \Sigma^{-1} K(y_T) \Sigma^{-1} \cdots \Sigma^{-1} K(y_{t+1})\Sigma^{-1} \right)^\top ; & \quad
     \tilde{b_t} =  \Sigma^{-1} K(y_t) \Sigma^{-1} \cdots K(y_1) \Sigma^{-1} \mu . \nonumber 
\end{eqnarray}
    \label{lemma:likelihood_decomposition}
\end{lemma}

We provide a detailed proof of this lemma in the supplementary material. The basic strategy is to fully expand $\widehat Pr(x_{1:T}) - Pr(x_{1:T}) $ after rewriting the estimated quantities as a sum of the true quantity plus an error term.  We then categorize each summand based on how many `$\Delta$' terms it has. There are three categories: terms with zero `$\Delta$' terms (i.e. the true likelihood $Pr(x_{1:T}) $), terms with only one `$\Delta$' (i.e. $(v+ \tilde{v})^\top \widehat{\Delta \mu} + \sum_{t=1}^T a_t^\top \widehat{\Delta K}(y_t) \tilde{a_t} - \sum_{t=0}^T b_t^\top \widehat{\Delta \Sigma} \tilde{b_t}$), and all remaining terms, which involve at least two `$\Delta$' quantities, and can be relegated to $\mathcal{O}_p (N^{-1})$. 

Lemma \ref{lemma:likelihood_decomposition} shows how the estimated error propagates to the likelihood approximation. We can leverage the fact that our moment estimators have a central limit theorem (CLT) property to obtain the desired results in Theroem \ref{theorem: CLT}.

We denote the outer product as $\otimes$, and  define a ``flattening'' operator $\mathcal{F} (\cdot) $ for both matrices and 3-way tensors. For matrix $A_{d \times d}$, $$\mathcal{F} (A) = [A^{(1, 1)}, A^{(1, 2)}, \cdots,  A^{(d, d)}]^\top;$$ For tensor $B_{d \times d \times d}$, $$\mathcal{F} (B) = [B^{(1,1,1)}, B^{(1,1,2)}, \cdots, B^{(1,1,d)}, B^{(1,2,1)}, B^{(1,2,2)}, \cdots, B^{(1,2,d)}, \cdots, B^{(1,d,d)}, \cdots, B^{(d, d ,d)}] ^\top$$

We now state and prove our main theorem.
\begin{theorem}
    \begin{eqnarray*}
    \sqrt{N} (\widehat Pr(x_{1:T}) - Pr(x_{1:T})) 
    & \xrightarrow[]{d} &
    N \left(  0,  \beta^\top 
        Cov \left( 
            \begin{bmatrix}
               Y_1 \\
               \mathcal{F} (Y_2 \otimes Y_1) \\
               \mathcal{F} (Y_3 \otimes Y_1 \otimes Y_2)
             \end{bmatrix}
         \right)
         \beta
    \right),
\end{eqnarray*}
where $$\beta = \left[ (v+ \tilde{v})^\top ; - \left( \sum_{t=0}^T \mathcal{F} (b_t \otimes \tilde{b}_t) \right)^\top ; \left( \sum_{t=1}^T \mathcal{F} (a_t \otimes \tilde{a}_t \otimes y_t) \right)^\top  \right]^\top $$
and $v, \tilde{v}, a_t, \tilde{a}_t, b_t, \tilde{b}_t$ are defined as in Lemma \ref{lemma:likelihood_decomposition}. 
\label{theorem: CLT}
\end{theorem}

\begin{proof*}[Theorem \ref{theorem: CLT}]
We flatten $\widehat{\Delta \Sigma}$ and $\widehat{\Delta K}$ as 
\begin{eqnarray}
\mathcal{F} (\widehat{\Delta \Sigma}) &=& [\widehat{\Delta \Sigma}^{(1, 1)}, \widehat{\Delta \Sigma}^{(1, 2)}, \cdots,  \widehat{\Delta \Sigma}^{(d, d)}]^\top ,\nonumber \\
\mathcal{F} (\widehat{\Delta K}) &=& [\widehat{\Delta K}^{(1, 1, 1)}, \widehat{\Delta K}^{(1, 1, 2)}, \cdots, \widehat{\Delta K}^{(d, d, d)}]^\top.
\nonumber\end{eqnarray}
Rewriting $a_t^\top \widehat{\Delta K}(y_t) \tilde{a_t}$ and $b_t^\top \widehat{\Delta \Sigma} \tilde{b_t} $ in Eq \ref{SimplifiedExpansion} as
\begin{eqnarray*}
a_t^\top \widehat{\Delta K}(y_t) \tilde{a_t} 
     & = & \mathcal{F} (a_t \otimes \tilde{a}_t \otimes y_t)^\top \cdot \mathcal{F} (\widehat{\Delta K}) ,  \\
    b_t^\top \widehat{\Delta \Sigma} \tilde{b_t} 
     & = & \mathcal{F} (b_t \otimes \tilde{b}_t)^\top \cdot \mathcal{F} (\widehat{\Delta \Sigma}) ,
\end{eqnarray*}
we have 
\begin{eqnarray*}
    & & \widehat Pr(x_{1:T}) - Pr(x_{1:T}) -  O_p (N^{-1}) \\
     & = & \left[ (v+ \tilde{v})^\top ; - \left( \sum_{t=0}^T \mathcal{F} (b_t \otimes \tilde{b}_t) \right)^\top ; \left( \sum_{t=1}^T \mathcal{F} (a_t \otimes \tilde{a}_t \otimes y_t) \right)^\top  \right]  \cdot
        \begin{bmatrix}
          \widehat{\Delta \mu} \\
          \mathcal{F} (\widehat{\Delta \Sigma}) \\
          \mathcal{F} (\widehat{\Delta K})
         \end{bmatrix} \\
    & = & \beta^\top \cdot \widehat{\Delta \theta}.
\end{eqnarray*}
 Since the CLT applies seperately to $\widehat{\Delta \mu} , \widehat{\Delta \Sigma} , \widehat{\Delta K} $, then
\begin{eqnarray*}
  \sqrt{N} \widehat{\Delta \theta} 
    & = & 
        \sqrt{N} \begin{bmatrix}
          \frac{1}{N} \sum_{i=1}^N Y_{i,1} - \mu \\
          \mathcal{F} (\frac{1}{N}  \sum_{i=1}^N Y_{i,2} \otimes Y_{i,1} -\Sigma) \\
          \mathcal{F} (\frac{1}{N} \sum_{i=1}^N Y_{i,3} \otimes Y_{i,1} \otimes Y_{i,2} - K)
         \end{bmatrix}
     \xrightarrow[]{d} 
    MVN \left(  \vec{0} , 
         Cov \left( 
            \begin{bmatrix}
              Y_1 \\
              \mathcal{F} (Y_2 \otimes Y_1) \\
              \mathcal{F} (Y_3 \otimes Y_1 \otimes Y_2)
             \end{bmatrix}
         \right) \right) .
\end{eqnarray*}
Therefore,
\begin{eqnarray*}
    \sqrt{N} (\widehat Pr(x_{1:T}) - Pr(x_{1:T})) 
    & \xrightarrow[]{d} &
    N \left(  0,  \beta^\top 
        Cov \left( 
            \begin{bmatrix}
              Y_1 \\
              \mathcal{F} (Y_2 \otimes Y_1) \\
              \mathcal{F} (Y_3 \otimes Y_1 \otimes Y_2)
             \end{bmatrix}
         \right)
         \beta
    \right) .
\end{eqnarray*}
\qed
\end{proof*}

\begin{figure}[!htb]
    \centering
    \includegraphics[width=0.25\linewidth]{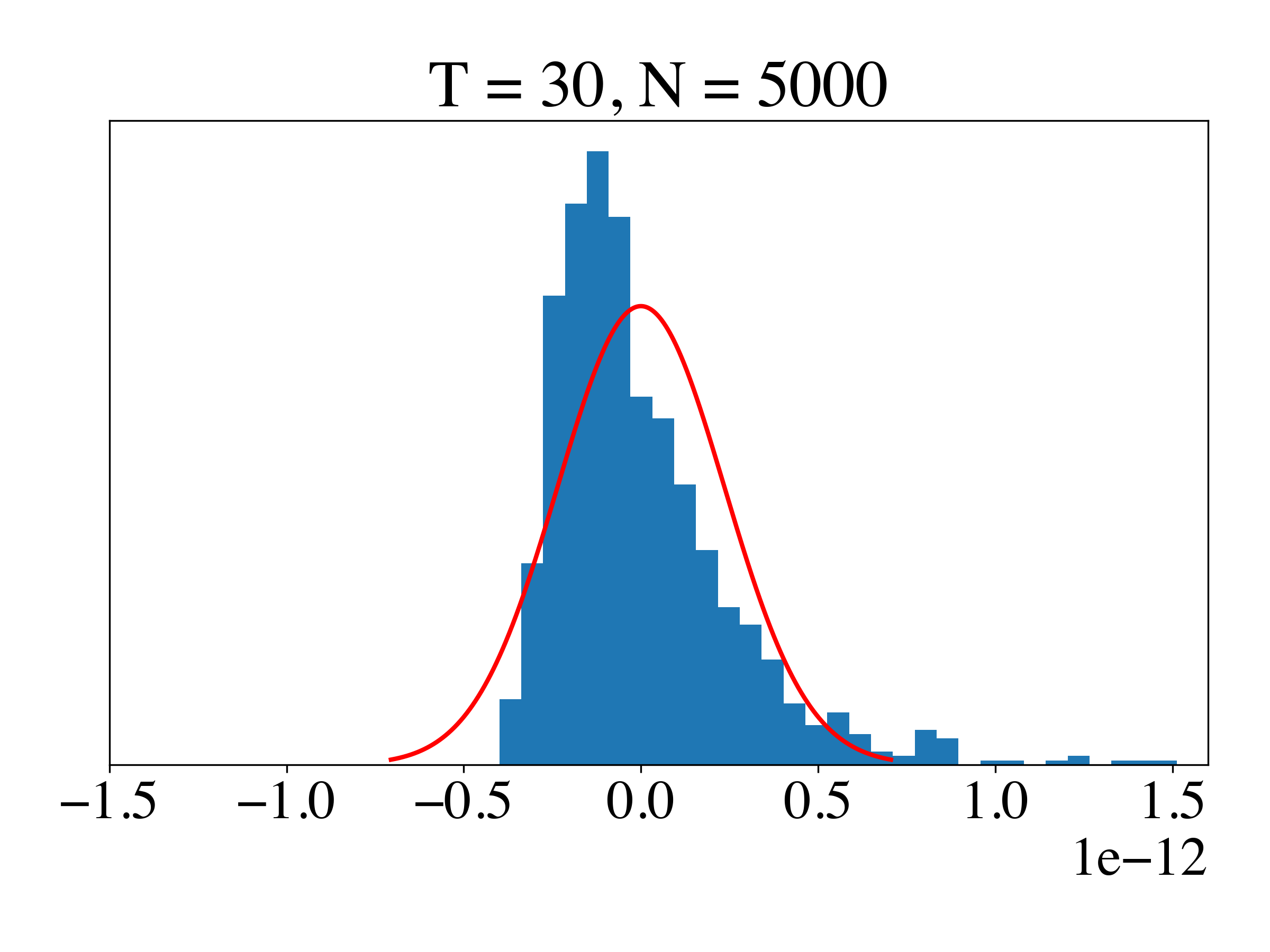}
    \hspace{-5mm}
    \includegraphics[width=0.25\linewidth]{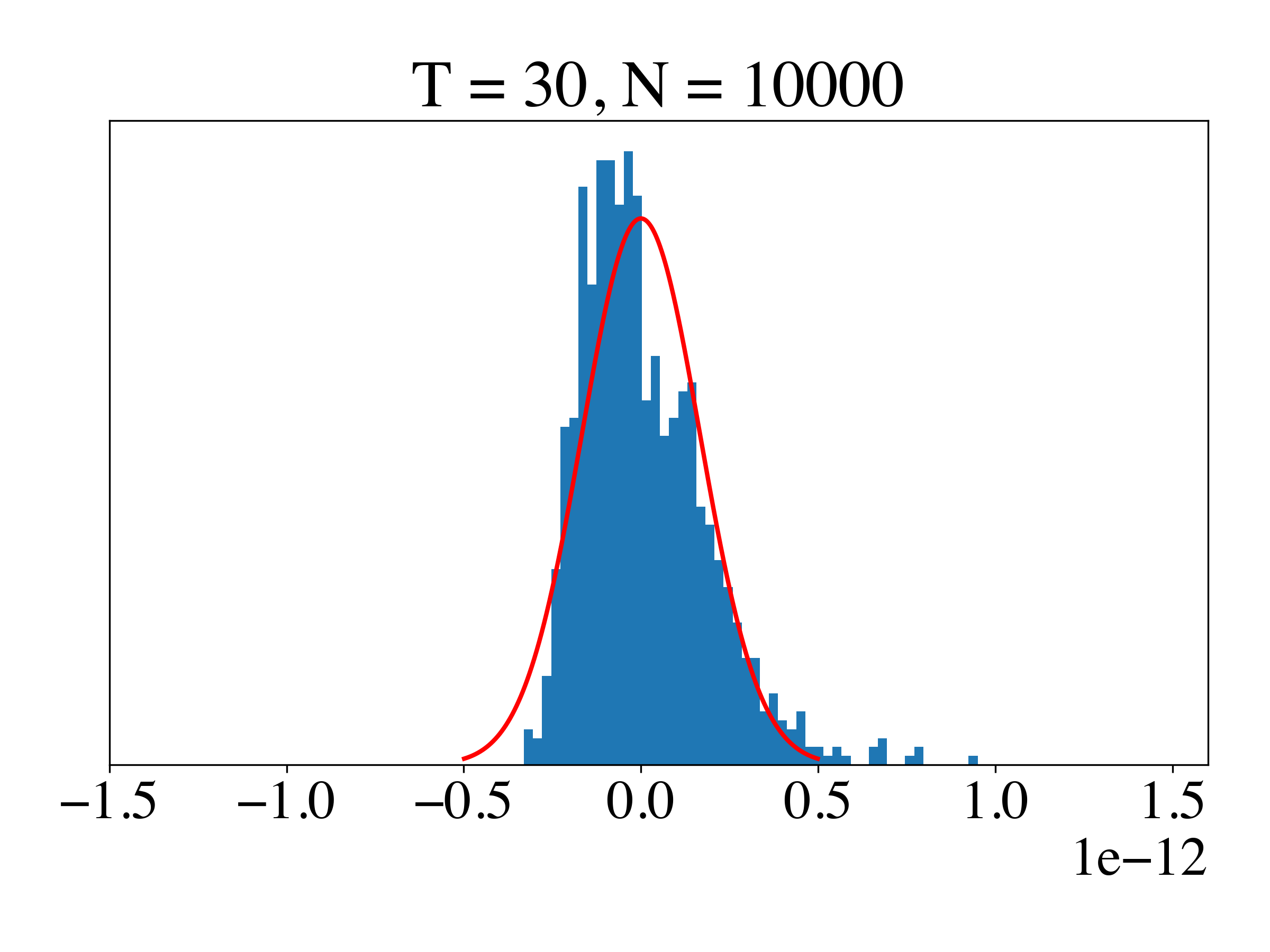}
    \hspace{-5mm}
    \includegraphics[width=0.25\linewidth]{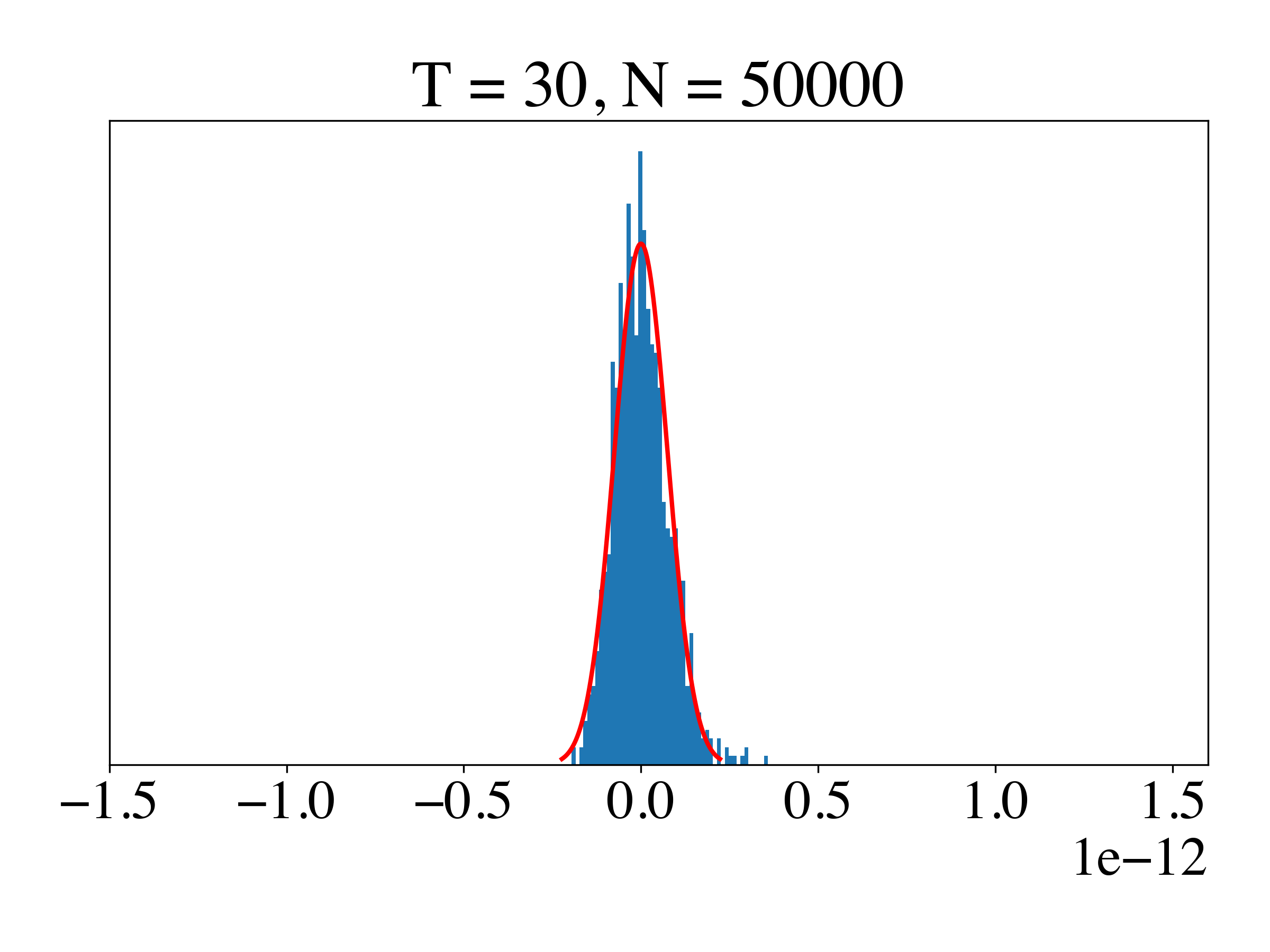}
    \hspace{-5mm}
    \includegraphics[width=0.25\linewidth]{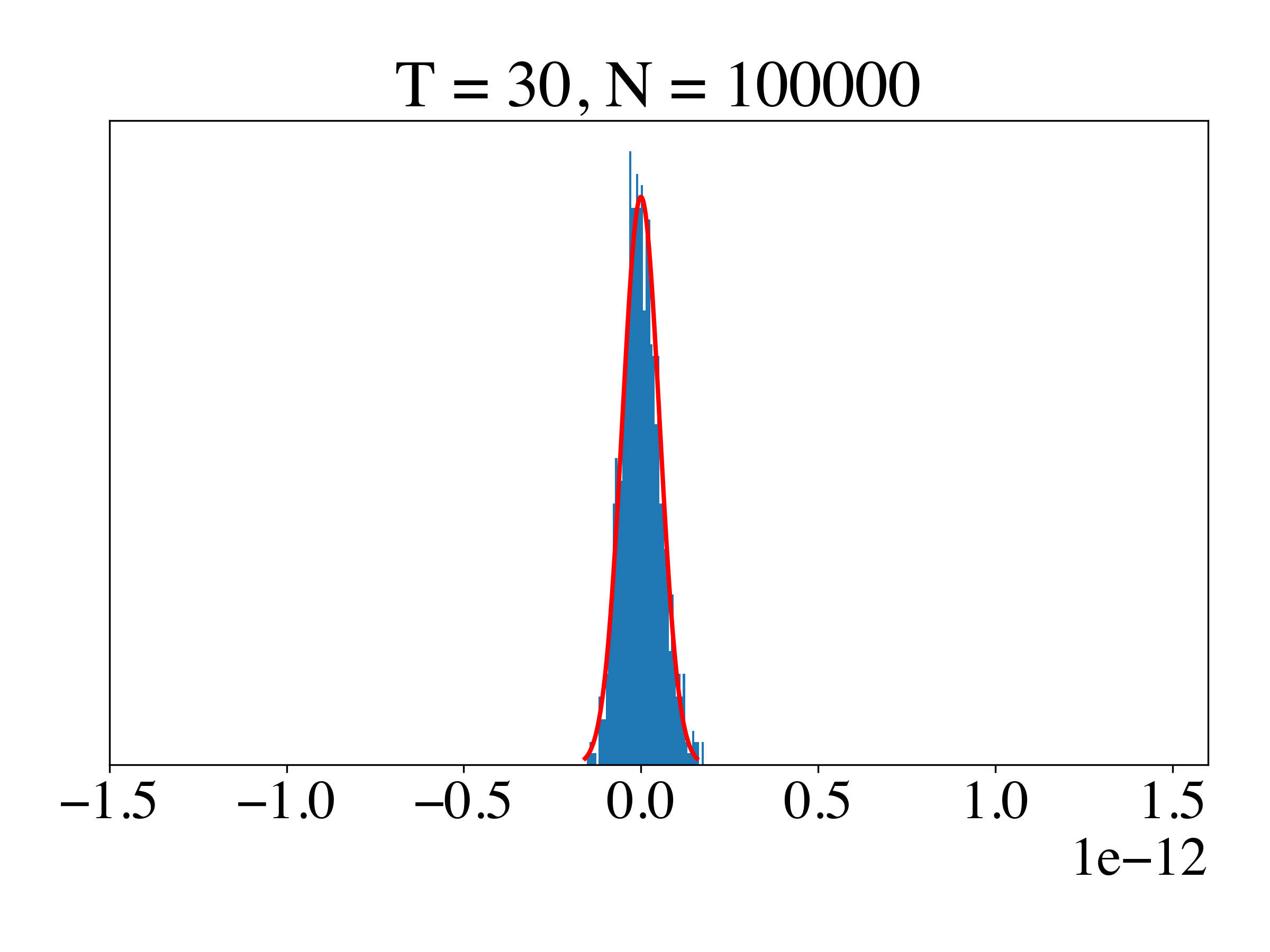}
    
    \includegraphics[width=0.25\linewidth]{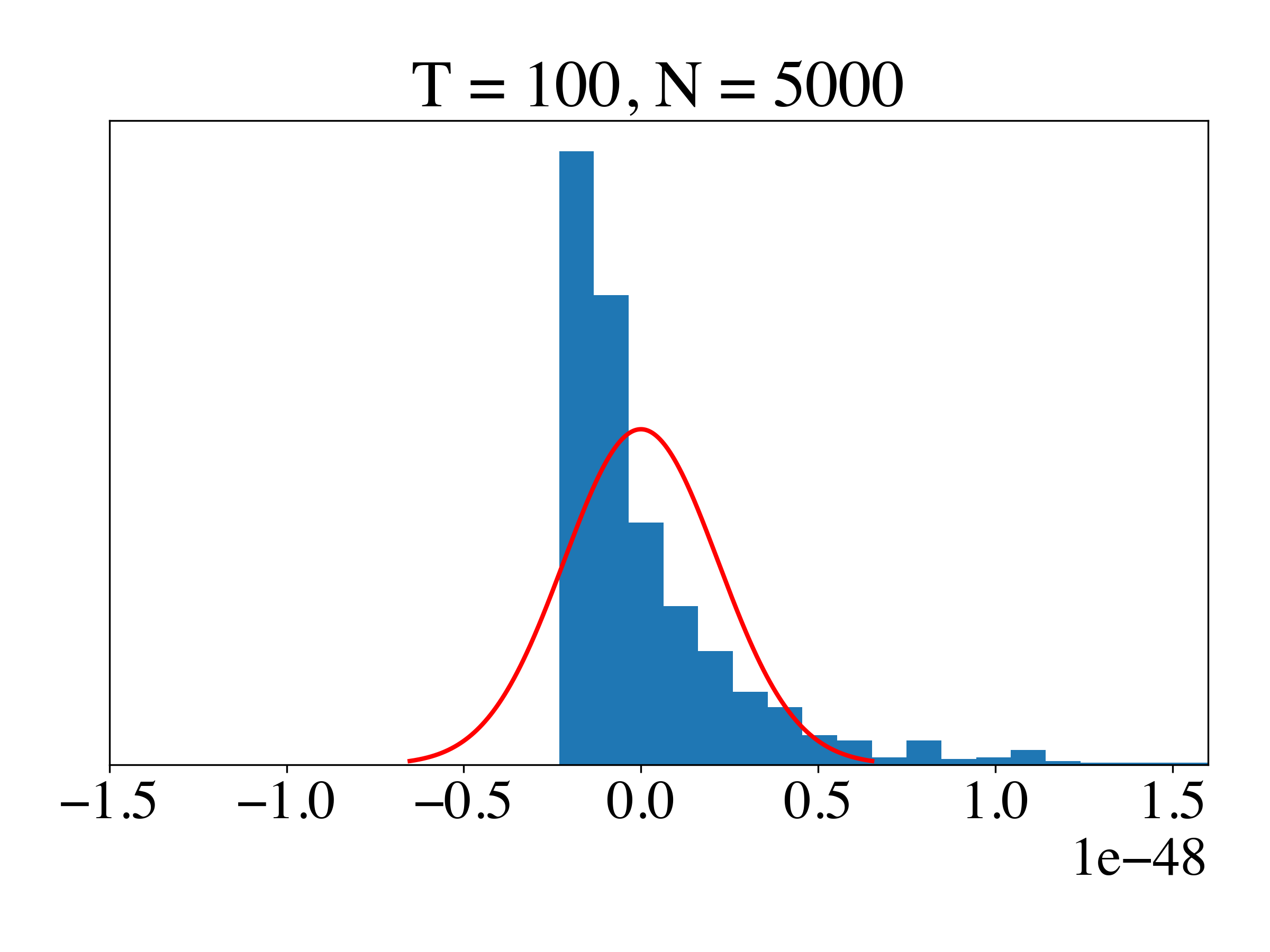}
    \hspace{-5mm}
    \includegraphics[width=0.25\linewidth]{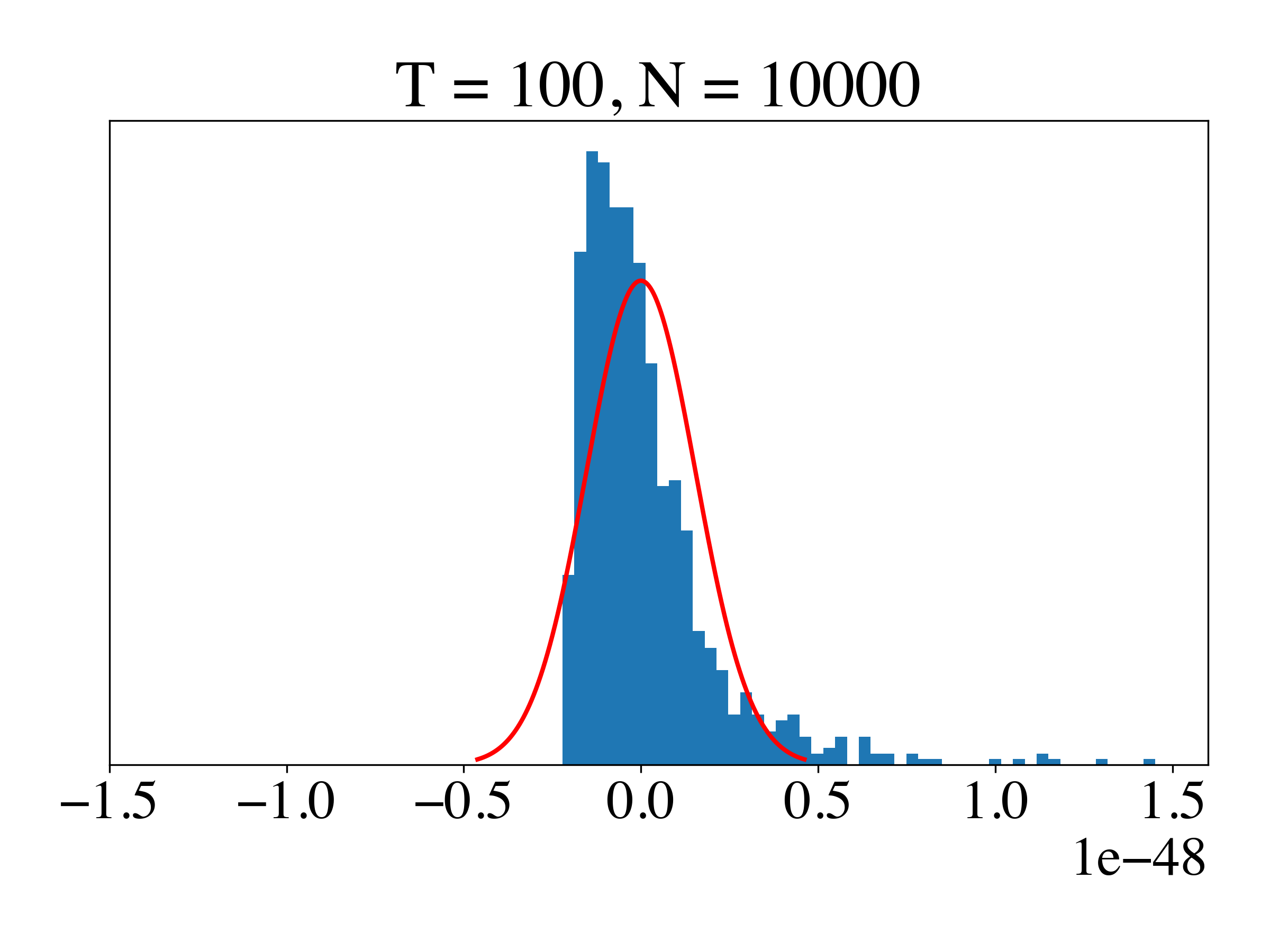}
    \hspace{-5mm}
    \includegraphics[width=0.25\linewidth]{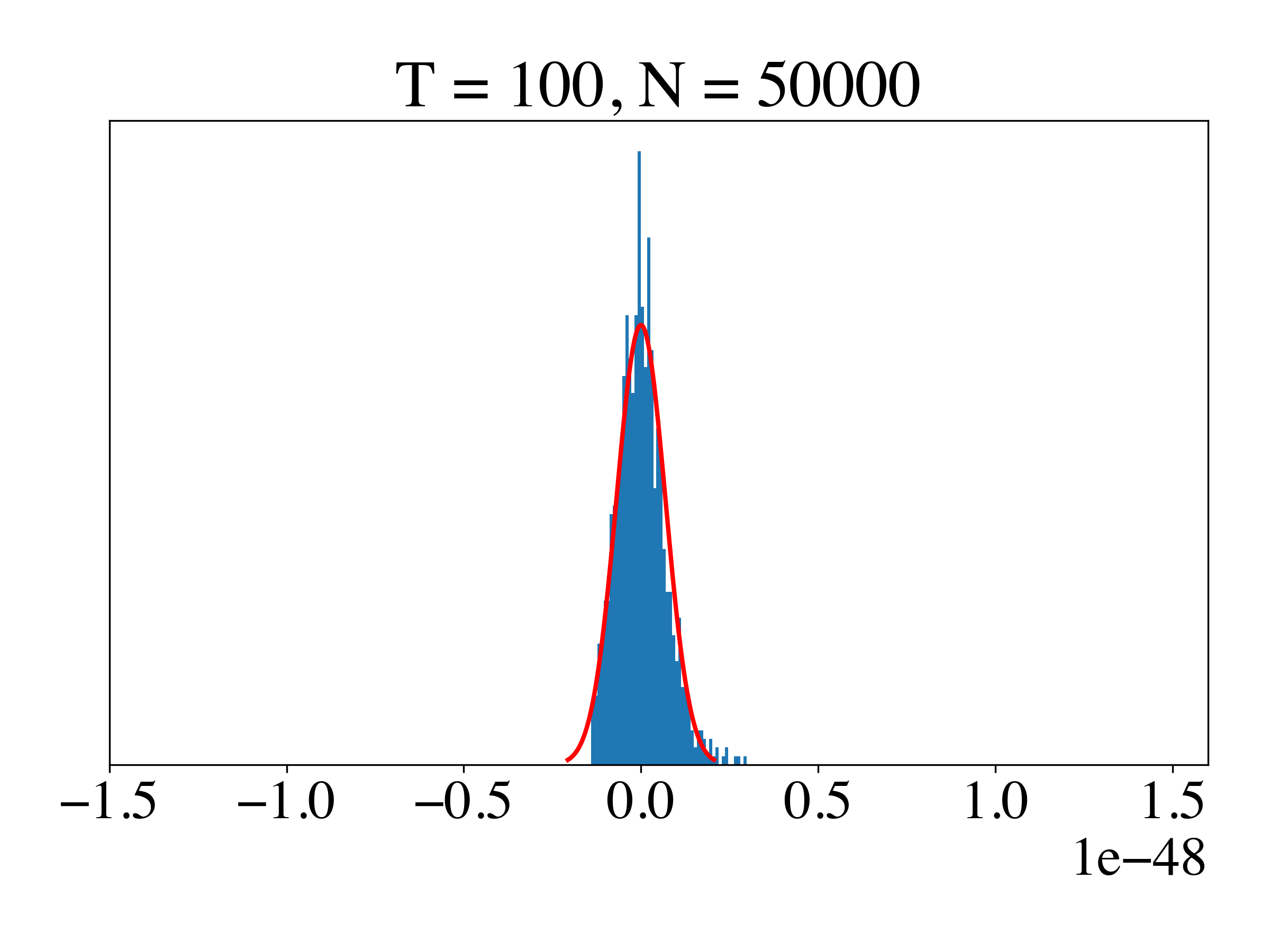}
    \hspace{-5mm}
    \includegraphics[width=0.25\linewidth]{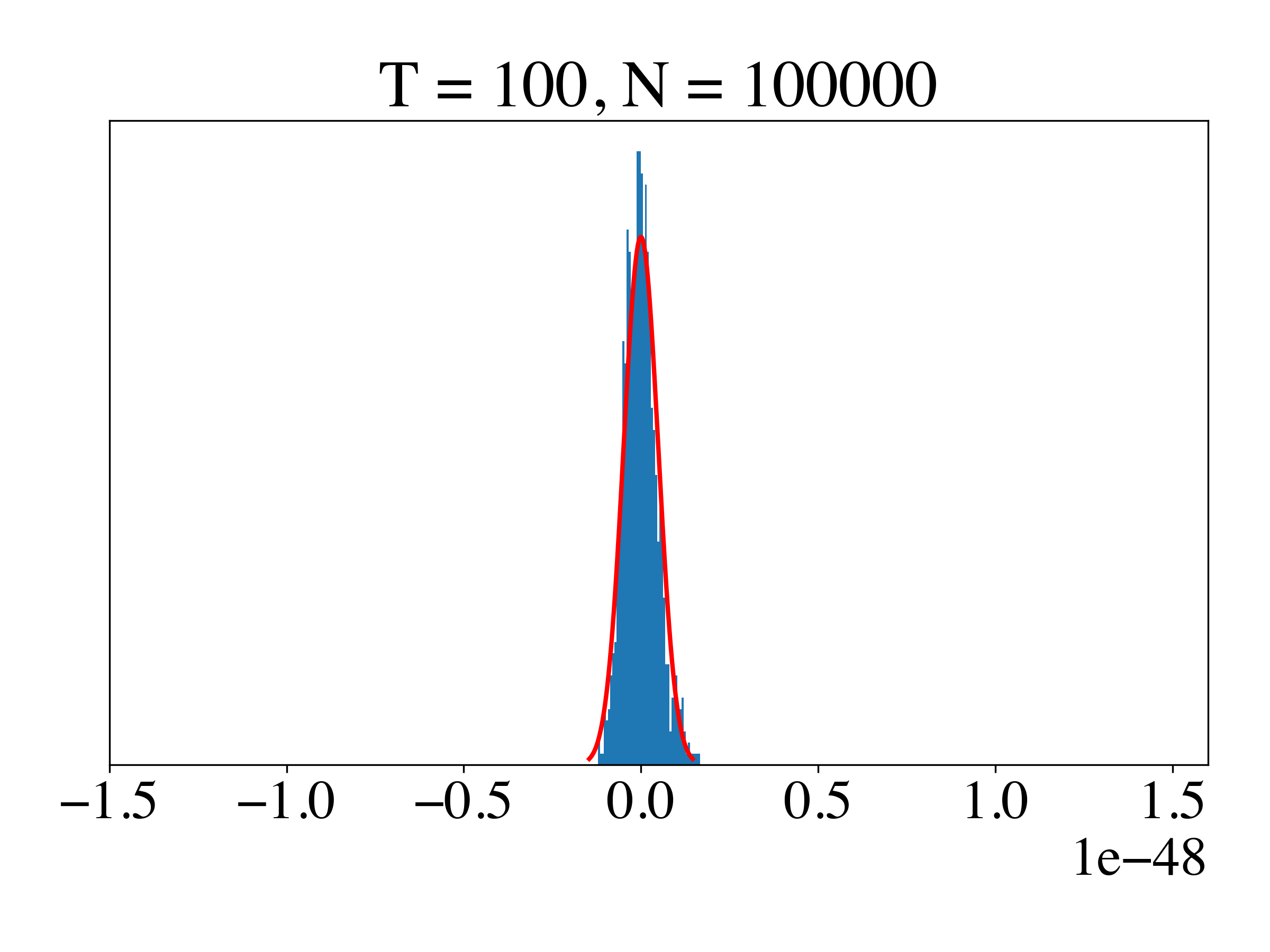}
    \caption{Convergence of SHMM estimation error to theoretical distribution. Empirical histograms of $\hat Pr (x_{1:T}) - Pr(x_{1:T})$ are shown for different training sizes $N$ and sequence lengths $T$, alongside the theoretical density derived from Theorem \ref{theorem: CLT}. Each subfigure corresponds to a different $N$. (1) As $N$ increases, the empirical distribution converges to the theoretical normal distribution. (2) Convergence is faster for smaller values of $T$.}
    \label{fig:Empirical pdf vs truth}
\end{figure}

\subsection{Experimental Validation of Theorem \ref{theorem: CLT}.}
We conducted a series of experiments to validate theorem \ref{theorem: CLT}. We simulated a target series $x_{1:T}$ with $x_i\in \mathbb{R}^3$ and $T = \{30, 100\}$ using a 3-state GHMM, where the initial probabilities and sticky transition probabilities are as described in Section \ref{subsubsec:experiment_setting}.  We set the discrete emission probability matrix to $[[0.8, 0.1, 0.1]^\top, [0.1, 0.8, 0.1]^\top, [0.1, 0.1, 0.8]^\top ]^\top $. We estimated parameters $\widehat{\mu}, \widehat{\Sigma}$, and $\widehat{K}$ using training samples generated from the same model as the target series. Specifically, $N$ i.i.d. samples $Y_1^{(\mu)}$ were used to estimate $\widehat{\mu}$, $N$ i.i.d. samples of $(Y_1^{(\Sigma)}, Y_2^{(\Sigma)})$ to estimate $\widehat{\Sigma}$, and $N$ i.i.d. samples of $(Y_1^{(K)}, Y_2^{(K)}, Y_3^{(K)})$ to estimate $\widehat{K}$.  We used training sets of size $N=5000,  10000, 50000, 100000$, replicating the experiment 1000 times for each $N$.  For each replication $r$, we estimated $\hat Pr (x_{1:T})^{(N, r)} $ and constructed a histogram of the estimation errors $\{ \hat Pr (x_{1:T})^{(N, r)} - Pr (x_{1:T})^{(N, r)} \}_r $.  We compared this histogram to the theoretical probability density function (pdf), which by Theorem \ref{theorem: CLT} should converge to a normal distribution as $N$ increases. 

\begin{figure} [ht]
    \hspace{-2.5cm}
	\begin{subfigure}[ht]{1.1\linewidth}
		\includegraphics[width=1.1\linewidth]{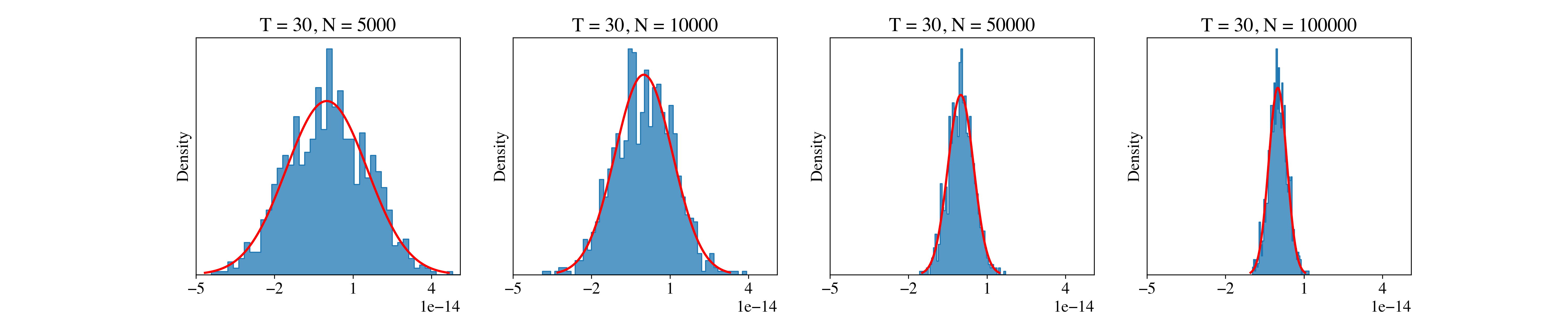}
		\caption{Likelihood estimation error from first-order first moment estimation.}
	\end{subfigure}
 
 \hspace{-2.5cm}
	\begin{subfigure}[ht]{1.1\linewidth}
		\includegraphics[width=1.1\linewidth]{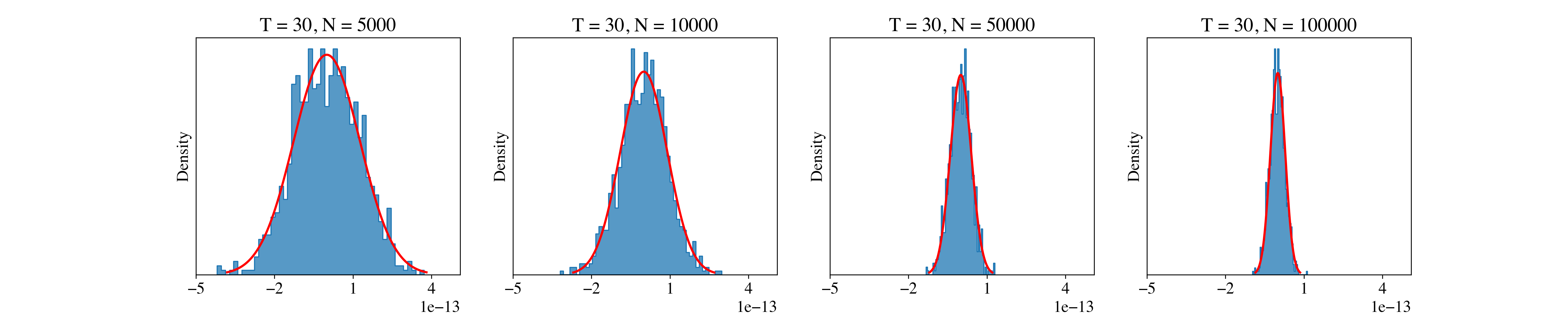}
		\caption{Likelihood estimation error from first-order second moment estimation.}
	\end{subfigure}
 
 \hspace{-2.5cm}
 \begin{subfigure}[ht]{1.1\linewidth}
		\includegraphics[width=1.1\linewidth]{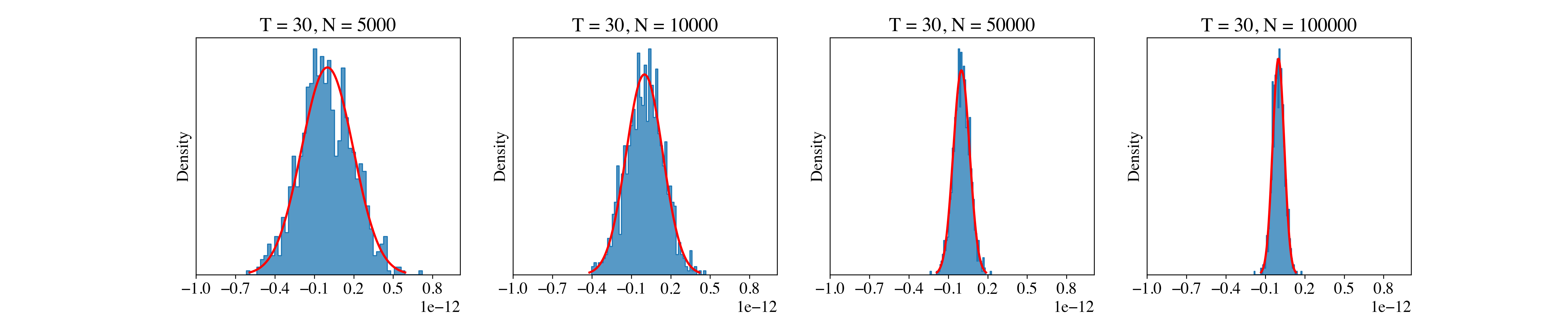}
		\caption{Likelihood estimation error from first-order third moment estimation.}
	\end{subfigure}
	
	\caption{Histogram of the first-order error from the first, second and third moment estimation error (i.e. $(v+ \tilde{v})^\top \widehat{\Delta \mu} $, $ \sum_{t=0}^T b_t^\top \widehat{\Delta \Sigma} \tilde{b_t}$, and $\sum_{t=1}^T a_t^\top \widehat{\Delta K}(y_t) \tilde{a_t}$) under different training sizes $N$ with fixed length $T=30$ vs. the theoretical pdf calculated based on Theorem \ref{theorem: CLT} (red line).
    Each subfigure is associated with a different $N$. As $N$ increases, the distribution converges to the theoretical normal distribution.}
	\label{fig.error_source_decomposition_his}
\end{figure}

\begin{figure} [ht]
    \hspace{-2.5cm}
	\begin{subfigure}[t]{1.1\linewidth}
		\includegraphics[width=1.1\linewidth]{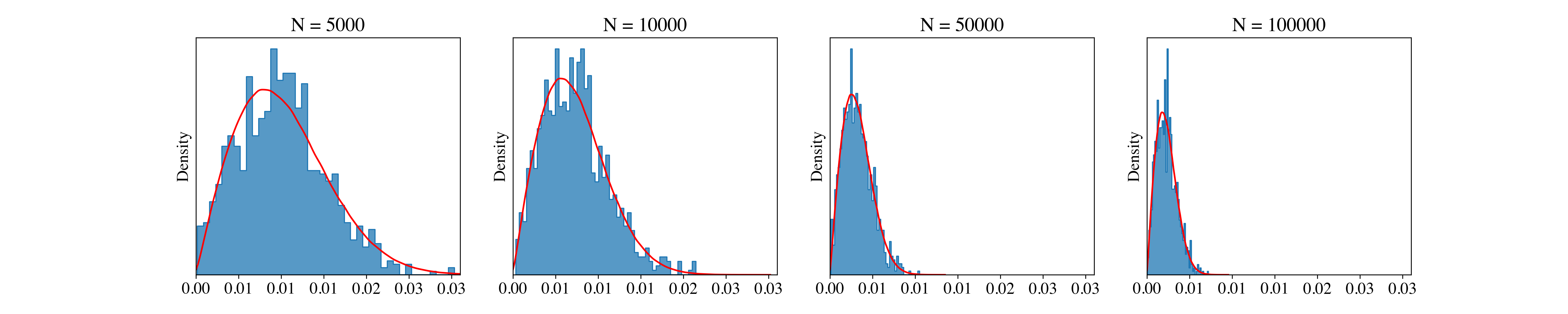}
		\caption{Frobenius norm of first moment estimation.}
	\end{subfigure}
 
 \hspace{-2.5cm}
	\begin{subfigure}[t]{1.1\linewidth}
		\includegraphics[width=1.1\linewidth]{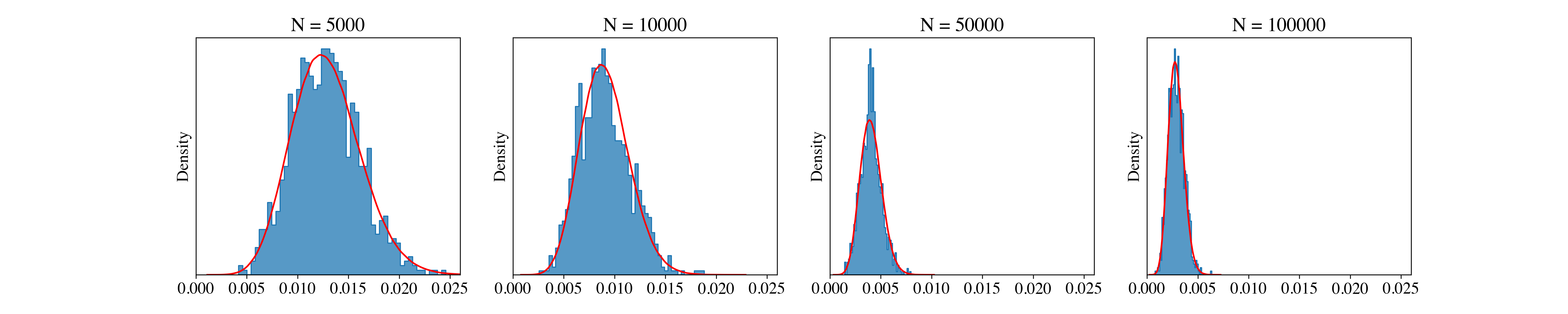}
		\caption{Frobenius norm of second moment estimation.}
	\end{subfigure}
 
 \hspace{-2.5cm}
 \begin{subfigure}[t]{1.1\linewidth}
		\includegraphics[width=1.1\linewidth]{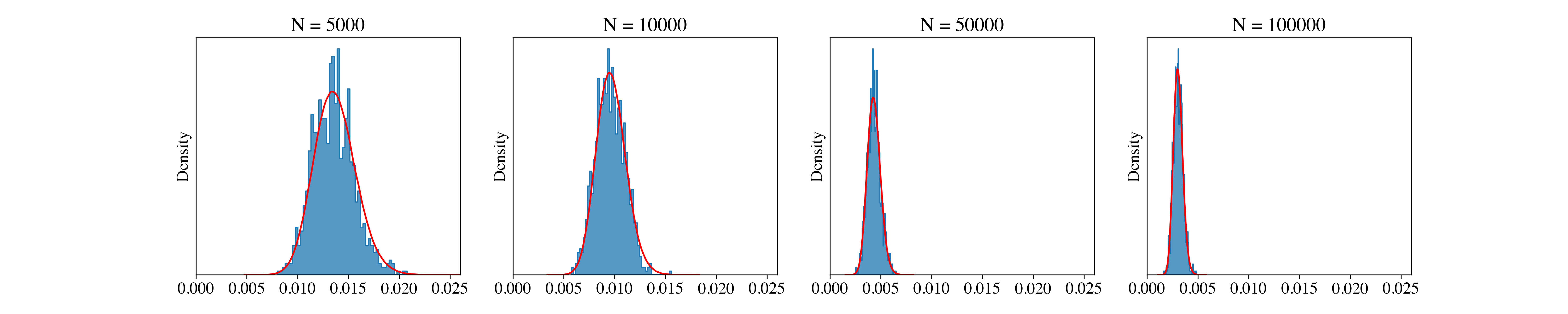}
		\caption{Frobenius norm of third moment estimation error}
	\end{subfigure}
	
	\caption{Histogram of the Frobenius norm of the first, second and third moment estimation error (i.e. $\mu$, $\Sigma$ and $K$) under different training size $N$ vs. the theoretical pdf (red line). Here the red line is the theoretical Chi-squared distribution. Each subfigure is associated with a different $N$. As $N$ increases, the distribution converges to the theoretical distribution.}
	\label{fig.norm_error_decomposition_his}
\end{figure}

In Figure \ref{fig:Empirical pdf vs truth} we see that, as $N$ increases, the distribution of the estimated likelihood converges to the normal distribution, and with a shorter length $T$, the error converges faster. We also analyzed the asymptotic behavior of the first-order estimation errors separately for the first moment, $ (v+ \tilde{v})^\top \widehat{\Delta \mu}$, second moment, $\sum_{t=0}^T b_t^\top \widehat{\Delta \Sigma} \tilde{b_t} $, and third moment, $\sum_{t=1}^T a_t^\top \widehat{\Delta K}(y_t) \tilde{a_t}$ (see Figure \ref{fig.error_source_decomposition_his}).  We found that the third moment estimation error has the largest effect, as shown in Table \ref{tab: simulation variance T = 30 100}. We further analyzed the asymptotic distribution of the Frobenius norm of the first, second and third moment estimation error. Figure \ref{fig.norm_error_decomposition_his} shows the empirical histogram and its corresponding theoretical pdf.

We note two sources of error in estimating the likelihood.  The first is the typical CLT-type error in parameter estimation, which decreases as $N$ increases. 
 The second the propagation of estimation errors during forward recursion.  To achieve a stable normal distribution given the second issue, $N$ must be much greater than $T$. In our simulation with $T=100$, we observe reasonable evidence of asymptotic normality at $N=10000$, but not at $N=5000$. When $N=T$ or is only slightly larger, the distribution is heavy-tailed, suggesting that regularization could be beneficial. 

For simplicity of presentation, we derive Theorem \ref{theorem: CLT} under the assumption that the output distribution is discrete. For continuous output, the CLT still holds with a proper kernel function $G(\cdot)$ as mentioned in section \ref{sec: SHMM_review}. 

\begin{table}[h]
    \begin{subtable}[h]{\textwidth}
        \centering
        \begin{tabular}{c|c|c|c}
            \hline
             Error source & Mathematical expression & Theoretical std & Variance explained ratio \\
             \hline
             1st moment estimation & $(v+ \tilde{v})^\top \widehat{\Delta \mu}$ & $1.09\times 10^{-12}  / \sqrt{N}$ & $0.43\%$ \\
             2nd moment estimation & $\sum_{t=0}^T b_t^\top \widehat{\Delta \Sigma} \tilde{b_t}$ & $8.98\times 10^{-12} / \sqrt{N}$ & $29.21\%$ \\
            3rd moment estimation & $\sum_{t=1}^T a_t^\top \widehat{\Delta K}(y_t) \tilde{a_t}$ & $1.39\times 10^{-11} / \sqrt{N}$ & $70.36\%$ \\
            \hline
        \end{tabular}
        \caption{$T=30$.}
        \label{tab: simulation variance T = 30}
    \end{subtable}

    \vspace{0.5cm}
\begin{subtable}[h]{\textwidth}
        \centering
        \begin{tabular}{c|c|c|c}
            \hline
             Error source & Mathematical expression & Theoretical std & Variance explained ratio \\
             \hline
             1st moment estimation & $(v+ \tilde{v})^\top \widehat{\Delta \mu}$ & $2.66\times 10^{-49}  / \sqrt{N}$ & $0.30\%$ \\
             2nd moment estimation & $\sum_{t=0}^T b_t^\top \widehat{\Delta \Sigma} \tilde{b_t}$ & $7.75\times 10^{-48} / \sqrt{N}$ & $25.37\%$ \\
            3rd moment estimation & $\sum_{t=1}^T a_t^\top \widehat{\Delta K}(y_t) \tilde{a_t}$ & $1.33\times 10^{-47} / \sqrt{N}$ & $74.59\%$ \\
            \hline
        \end{tabular}
        \caption{$T=100$.}
        \label{tab: simulation variance T = 100}
    \end{subtable}
     \caption{Theoretical variance for first, second, third moment estimation errors based on simulated data with different $T$. }
    \label{tab: simulation variance T = 30 100}
\end{table}



\section{Projected SHMM} \label{sec:regularized SHMM}
\subsection{Motivation for Adding Projection}

The Baum-Welch algorithm \citep{baum1970maximization} predicts belief probabilities, or weights, for each underlying hidden state. Let $\hat w_t$ denote the predicted weight vector at time $t$. The prediction can be expressed as a weighted combination of cluster means: $\hat{y}_t = M \hat{w}_t $ where $||\hat{w}_t||_1 = 1$. In the Baum-Welch algorithm, these weights are explicitly guaranteed to be non-negative and sum to 1 during forward propagation, which is consistent with their interpretation as probabilities.

However, SHMM does not directly estimate the belief probabilities, and these constraints are not explicitly enforced in spectral estimation. Consequently, SHMM can sometimes produce predictions that are far from the polyhedron spanned by the cluster means. To address this issue, we propose the projected SHMM, where projection serves to regularize the predictions, constraining them within a reasonable range.

This is particularly important when $N$ is not sufficiently large, and extreme deviations of the estimated likelihood from the true likelihood can occur when error is propagated over time. Regularization can help stabilize the performance of estimation of the likelihood by limiting this propagation of error.

\subsection{Projection Methods for SHMM}

We propose two projection methods for SHMM: projection-onto-polyhedron and projection-onto-simplex.  Projection-onto-polyhedron directly addresses the motivation for using projections in SHMM but is computationally expensive. As a computationally efficient alternative, we propose projection-onto-simplex.

\subsubsection{Projection-onto-Polyhedron}
Projection-onto-polyhedron SHMM first generates prediction $\hat{y}_t^{(SHMM)}$ through the standard SHMM and then projects it onto the polyhedron with vertices $\widehat{M}$.  
Mathematically, we replace the recursive forecasting in Eq \ref{eq:recursive_prediction_rodu} with
\begin{eqnarray}
    \hat{y}_t^{(SHMM)} & = & \frac{\hat C (y_{t-1}) \hat{y}_{t-1} }{\hat c^\top_{\infty} \hat C (y_{t-1}) \hat{y}_{t-1} } ; \nonumber\\
    \hat{y}_t & = & \operatorname*{\arg \min}_{y \in Poly (\widehat{M}) } d (y, \hat{y}_t^{(SHMM)}) \label{eq:polyhedron},
\end{eqnarray}
where $d ( \cdot , \cdot ) $ is a distance function (e.g., Euclidean distance), and $$ Poly (\widehat{M}) = \{ y = \widehat{M} w | w \ is \ on \ the \ simplex  \}$$ is the polyhedron with vertices $\widehat{M}$.  This results in a convex optimization problem if the distance is convex.  While solvable using standard methods such as the Newton-Raphson algorithm \citep{boyd2004convex_optimization_textbook} with log-barrier methods \citep{frisch1955logbarrier}, to the best of our knowledge finding an \emph{efficient} solution for projection-onto-polyhedron is challenging. 
 The optimization must be performed at every time step, creating a trade-off between accuracy and computational speed.  Because speed over the Baum-Welch algorithm is one of the distinct advantages of SHMM, any modification should not result in a substantial increase in computational time. 


\subsubsection{Projection-onto-Simplex}
\begin{figure}[ht]
	\centering
	\hspace{-0.8cm}
	\begin{subfigure}[t]{0.45\linewidth}
	    \centering
		\includegraphics[width=2in]{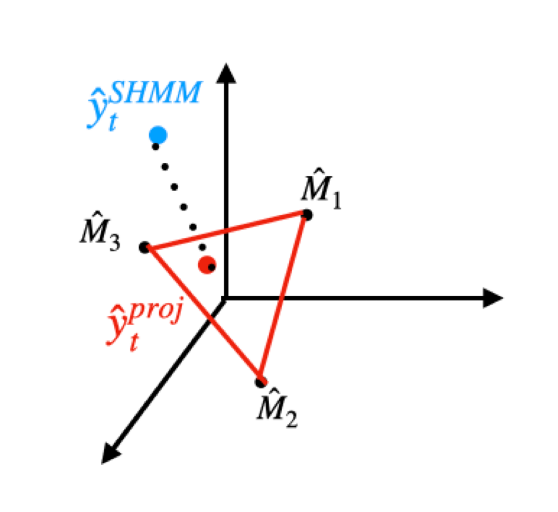}
 		\caption{Projection-onto-polyhedron.}
		\label{fig:polyhedron}
	\end{subfigure}
	\begin{subfigure}[t]{0.45\linewidth}
	    \centering
		\includegraphics[width=2in]{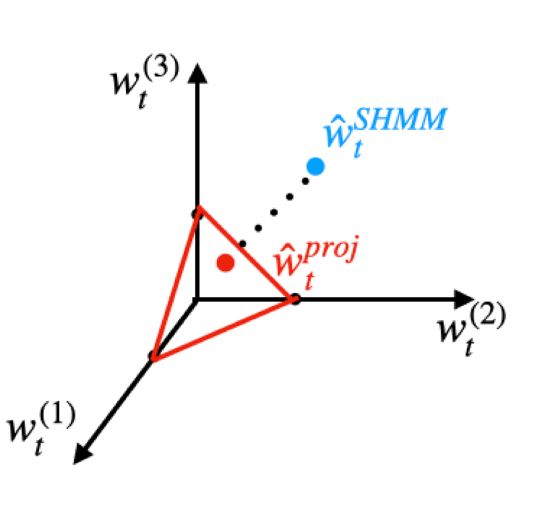}
 		\caption{Projection-onto-simplex.}
		\label{fig:simplex}
	\end{subfigure}
	\caption{The left figure shows the projection-onto-polyhedron step, and the right shows projection-onto-simplex. In both methods, we project the predicted values (blue points) into the constrained regions (defined with a red boundary), a polyhedron (left) or simplex (right). }
	\label{fig:polyhedron_simplex}
\end{figure}

To improve computational efficiency, we propose projection-onto-simplex. To avoid projection onto a polyhedron, we leverage the fact that $\hat{y}_t = \widehat{M} \hat{w}_t $ and optimize over $\hat{w}_t$ on the simplex:
\begin{eqnarray}
    \hat{w}_t & = & \operatorname*{\arg \min}_{w \in Simplex } || w - \widehat{M}^{-1} \hat{y}_t^{(SHMM)} ||_2^2 .
    \label{eq:proj_simplex_loss}
\end{eqnarray}
This allows us to calculate the projection with time complexity $\mathcal{O} (d \log (d ))$ \citep{wang2013proj_onto_simplex}.  This approach, while not equivalent to the solution from the projection-onto-polyhedron--$d(a, b) \neq d(A a, A b) $ in general--ensures predictions are constrained to the same polyhedron. Importantly, it can be solved using a closed-form solution (Algorithm \ref{alg:projection-onto-simplex}), avoiding iterative optimization and yielding faster estimation. Figure \ref{fig:polyhedron_simplex} illustrates the projection-onto-polyhedron and projection-onto-simplex methods.

\begin{algorithm}[!tb]
    \SetAlgoLined
    \SetKwInOut{Input}{Input}
    \SetKwInOut{Output}{Output}

    \Input{$u = [u_1, u_2, \cdots, u_d]^\top$} 
    Sort $u$ into $z$: $z_1 \geq z_2 \cdots \geq z_d$\;
    Find $\rho = max\{ 1 \leq i \leq d: z_i + \frac{1}{i} (1 - \sum_{j=1}^i z_j) > 0 \}$\;
    Define $\lambda = \frac{1}{\rho} (1 - \sum_{j=1}^{\rho} z_j)$\;
    Solve $u^{(proj)}$, s.t. $u_i^{(proj)} = max(u_i + \lambda, 0)$, $i = 1, \cdots, d$\;
    \Output{$u^{(proj)}$}
    
    \caption{Projection-onto-simplex \citep{wang2013proj_onto_simplex}.}
    \label{alg:projection-onto-simplex}
\end{algorithm}

The full projected SHMM algorithm is shown in Algorithm \ref{alg:PSHMM algo}. In Algorithm \ref{alg:PSHMM algo}, Steps 1-3 are identical to the standard SHMM. Steps 4-5 estimate $\hat{M}$ by Gaussian Mixture Models (GMM) \citep{mclachlan1988gmm}, calculate the weight processes $\{ w_t \} $, and apply SHMM on the weight process. Step 6 applies projection-onto-simplex on the recursive forecasting. Step 7 projects the data back into the original space. 

\begin{algorithm}[!tb]
    \SetAlgoLined
    \SetKwInOut{Input}{Input}
    \SetKwInOut{Output}{Output}
   
    \Input{$\{x_t\}$, where $t = 1, \cdots, T$}
    \Output{$\hat x_{T+1}$}

    Step 1: Compute $\hat E[x_{t+1} \otimes x_{t}] = \frac{1}{T-2} \sum_{i = 1} ^ {T-2} x_{t+1} x_t^\top$\;
    
    Step 2: Obtain $\hat U$ by extracting the first $k$ left eigenvectors of $\hat E[x_{t+1} \otimes x_{t}]$\;

    Step 3: Reduce dimensionality $ y_t = \hat U ^\top x_t$\;

    Step 4: Estimate cluster mean by GMM, and obtain $\hat M$, where each column is the mean vector of each cluster. Then the weight vector is $w_t = \hat M ^{-1} y_t$ for $t = 1, \cdots, T$\;

    Step 5: Calculate $\hat \mu = \frac{1}{T} \sum_{t=1}^\top w_t$, $\hat \Sigma = \frac{1}{T-1} \sum_{t=1}^{T-1} w_{t+1}w_t^\top$, and $\hat K = \frac{1}{T-2} \sum_{t = 1}^{T-2} w_{t+2} \otimes w_t \otimes w_{t+1}$. Set $\hat c_1 = \hat \mu$, $\hat c_{\infty}^\top = c_1^\top \hat \Sigma ^{-1}$, and $\hat C(w_t) = \hat K(w_t) \hat \Sigma ^{-1}$\;
    
    Step 6: Recursive prediction with projection-onto-simplex $\hat w_t = Proj \left( \frac{\hat C (w_{t-1}) \hat{w}_{t-1} }{\hat c^\top_{\infty} \hat C (w_{t-1}) \hat{w}_{t-1} } \right) $ for $t=2, \cdots, T+1$ where $Proj(a) =\operatorname*{\arg \min}_{w \in Simplex } || w - a ||_2^2$ can be solved by Algorithm \ref{alg:projection-onto-simplex}, and set $\hat{y}_1 = \hat c_1$\;

    Step 7: $\hat x_{T+1} = \hat U \hat y_{T+1} = \hat U \hat M \hat w_{T+1}$\;
    
    \caption{Projection-onto-simplex SHMM.}
    \label{alg:PSHMM algo}
\end{algorithm}


\subsubsection{Bias-variance tradeoff}\label{bias_variance_tradeoff}
In PSHMM, we leverage GMM to provide projection boundaries, which would introduce bias, since the hidden state means estimated by GMM are biased when we ignore time dependency information. In addition, either projection method-- 'projection-onto-polyhedron' or 'projection-onto-simplex'-- can introduce bias since they are not necessarily an orthogonal projection due to optimization constraints. However, adding such a projection will largely reduce the variance. That is, there is a bias-variance tradeoff.

\subsection{Considerations for PSHMM Implementation}
\label{subsec: pshmm hyperparameter choice}

This section addresses two key aspects of PSHMM implementation: choosing the dimensionality of the projection space and computing the projection matrix $U$.  

\subsubsection{Choice of projection space dimensionality}
\label{sec: discussion on d}

The hyperparameter $d$ determines the dimensionality of the projection space. Theoretically, $d$ should equal the number of states in the HMM. Our simulations confirm that when $d$ matches the true number of underlying states, estimation and prediction performance improves. However, in practice, the number of hidden states is often unknown. We recommend either choosing $d$ based on prior knowledge or tuning it empirically when strong prior beliefs are absent.

\subsubsection{Computation of projection matrix $U$}

The projection matrix $U$ is typically constructed using the first $d$ left singular vectors from the singular value decomposition (SVD) \citep{eckart1936svd} of the bigram covariance matrix $\widehat{\Sigma} = \widehat{\mathrm{E}} [X_2 \otimes X_1] $. This approach encodes transition information while eliminating in-cluster covariance structure. Alternatively, $U$ could be estimated through an SVD of $\widehat{\mathrm{E}} [X_1 \otimes X_1] $, which would incorporate both covariance structure and cluster mean information. We generally recommend using the bigram matrix approach.



For extremely high-dimensional data, we propose a fast approximation algorithm based on randomized SVD \citep{halko2011random_svd_complexity}. This method avoids computing the full covariance matrix $\hat E[x_{t+1} \otimes x_{t}]$, reducing time complexity from $\mathcal{O} (T p^2 + p^3 )$ to $\mathcal{O} (pT \log (d) + (p + T) d^2 )$ for $p \gg d$, where $T$ and $p$ are the sample size and data dimensionality, respectively.  The algorithm proceeds as follows:

\begin{enumerate}
    \item Noting that $ \hat E[x_{t+1} \otimes x_{t}] = \frac{1}{T-2} \sum_{i = 1} ^ {T-2} x_{t+1}x_t^\top = \frac{1}{T-2} X_2^\top X_1 $, where $ X_2 = [x_2 , \cdots, x_T]^\top$ and $ X_1 = [x_1 , \cdots, x_{T-1}]^\top$, perform a randomized SVD of $X_1$ and $X_2$ separately to obtain two rank-$\tilde{d}$ decompositions with $d \leq \tilde{d} \ll p$: $X_1 \approx U_1 \Sigma_1 V_1^\top $, $X_2 \approx U_2 \Sigma_2 V_2^\top $.

    \hspace{.5cm} Now $ X_2^\top X_1 \approx V_2 (\Sigma_2 U_2^\top U_1 \Sigma_1 ) V_1^\top $, where $(\Sigma_2 U_2^\top U_1 \Sigma_1 )$ has dimension $\tilde{d} \times \tilde{d}$.
    \item Perform an SVD on $(\Sigma_2 U_2^\top U_1 \Sigma_1 )$ to get $\Sigma_2 U_2^\top U_1 \Sigma_1 = \tilde{U} \tilde{\Sigma} \tilde{V}^\top$.
    \item We can reconstruct $\hat E[x_{t+1} \otimes x_{t}] \approx (V_2 \tilde{U}) (\frac{1}{T-2} \tilde{\Sigma}) (V_1 \tilde{V})^\top$.
\end{enumerate}

$V_2 \tilde{U}$ and $V_1 \tilde{V}$ are orthonormal matrices and $\frac{1}{T-2} \tilde{\Sigma}$ is a diagonal matrix, so this is the rank-$\tilde{d}$ SVD of $\hat E[x_{t+1} \otimes x_{t}]$. The first $d$ vectors of $V_2 \tilde{U}$ are an estimate of the $\hat U$ matrix we compute in Step 1 and 2 in Algorithm \ref{alg:PSHMM algo}

\section{Online Learning} \label{sec: online learning}

Online learning is a natural extension for SHMMs, given their computational efficiency advantages. This section introduces online learning approaches for both SHMM and PSHMM, addressing scenarios where traditional batch learning may be unsuitable.  For example, in quantitative trading, markets often exhibit a regime-switching phenomenon.  Thus, it is likely that the observed returns for some financial products are nonstationary. In high frequency trading, especially in second-level or minute-level trading, the delay from frequent offline re-training of the statistical model could impact the strategy and trading speed \citep{lahmiri2021deep}. 

\subsection{Online Learning of SHMM and PSHMM}

We illustrate an online learning strategy that allows for adaptive parameter estimation.  Let $\hat{\mu}, \hat{\Sigma}$, and $\hat{K}$, be the estimated moments based on $T$ data points.  Upon receiving new data $Y_{T + 1} $, we update our moments recursively:


\begin{eqnarray}
    \hat{\mu} & \xleftarrow{} & \frac{T \cdot \hat{\mu} + Y_{T + 1}}{T + 1} ; \nonumber\\
    \hat{\Sigma} & \xleftarrow{} & \frac{(T-1) \cdot \hat{\Sigma} + Y_{T+1} \otimes Y_{T}}{T} ; \nonumber\\
    \hat{K} & \xleftarrow{} & \frac{(T-2) \cdot \hat{K} + Y_{T+1} \otimes Y_{T-1} \otimes Y_{T}}{T-1}  ; \nonumber\\
    T & \xleftarrow{} & T + 1 .
    \label{eq:online_update_moment_nodecay}
\end{eqnarray}
The update strategy applies to both SHMM and PSHMM.  Algorithm \ref{alg:online PSHMM algo} provides pseudo-code for online learning in PSHMM.

\begin{algorithm}[!tb]
    \SetAlgoLined
    \SetKwInOut{Input}{Input}
    \SetKwInOut{Output}{Output}
   
    \Input{$\{x_t \}_{t = 1, \cdots, T} $, the warm-up length $T_{warmup}$ }
    \Output{$ \{ \hat x_t \}_{t=T_{warmup} + 1}^{T+1} $ yielded sequentially. }

    Step 1: Compute $\hat E[x_{t+1} \otimes x_{t}]^{(warmup)} = \frac{1}{T_{warmup}-2} \sum_{i = 1} ^ {T_{warmup}-2} x_{t+1} x_t^\top$\;
    
    Step 2: Obtain $\hat U$ by extracting the first $k$ left eigenvectors of $\hat E[x_{t+1} \otimes x_{t}]^{(warmup)}$\;

    Step 3: Reduce dimensionality $y_t = \hat U ^\top x_t$ for $t = 1, \cdots, T_{warmup} $\;

    Step 4: Estimate cluster mean by GMM by data $ \{ \hat y_t \}_{t=1}^{T_{warmup}}$, and obtain $\hat M$, where each column is the mean vector of each cluster. Then the weight vector is $w_t = \hat M ^{-1} y_t$ for $t = 1, \cdots, T_{warmup}$\;

    Step 5: Calculate $\hat \mu$, $\hat \Sigma$, $\hat K$, $\hat c_1$, $\hat c_{\infty}^\top$ and $\hat C(\cdot)$ as described in Step 5 of Algorithm \ref{alg:PSHMM algo}\;
     
    Step 6: Recursive prediction with projection-onto-simplex $\hat{w}_t$ for $t=1, \cdots, T_{warmup} +1$ as described in Step 6 of Algorithm \ref{alg:PSHMM algo}. Yield $ \hat x_{T_{warmup} +1} = \hat U \hat M \hat w_{T_{warmup} +1} $\;
    
    Step 7 (online learning and prediction): 
    
    \For{$t \leftarrow T_{warmup} +1$ \KwTo $T $}{
        
        $w_t = \hat M ^{-1} \hat U ^\top x_t $\;
        
        Update $\hat \mu$, $\hat \Sigma$ and $\hat K$ according to Eq \ref{eq:online_update_moment_nodecay}\;
        
        Predict $\hat w_{t + 1}$ by $\hat w_{t + 1} = Proj \left( \frac{\hat C (w_{t}) \hat{w}_{t} }{\hat c^\top_{\infty} \hat C (w_{t}) \hat{w}_{t} } \right) $, where $ \hat C (\cdot) $ and $\hat c^\top_{\infty}$ are based on updated $\hat \mu$, $\hat \Sigma$ and $\hat K$, and $Proj (\cdot)$ is solved by Algorithm \ref{alg:projection-onto-simplex}\;

        Yield $ \hat x_{t +1} = \hat U \hat M \hat w_{t + 1} $\;
    }
    \caption{Online learning projection-onto-simplex SHMM.}
    \label{alg:online PSHMM algo}
\end{algorithm}

\paragraph{Updating the GMM used in PSHMM}
For PSHMM, updating the Gaussian Mixture Model (GMM) requires careful consideration. We recommend updating the GMM without changing cluster membership. For example, classify a new input into a particular cluster and update that cluster's mean and covariance. We advise against re-estimating the GMM, which could add or remove clusters, as this would alter the relationship between $y$'s dimensionality and the number of hidden states. In practice, we find PSHMM performs well even without GMM updates.


\subsection{Online Learning of SHMM Class with Forgetfulness}
To handle nonstationary data, we introduce a forgetting mechanism in online parameter estimation. This approach requires specifying a decay factor $\gamma$, which determines the rate at which old information is discarded. The updating rule becomes:
\begin{eqnarray}
    \hat{\mu} & \xleftarrow{} & \frac{(1 - \gamma) \Tilde{T} \hat{\mu} + Y_{T + 1}}{(1 - \gamma) \Tilde{T} + 1} ; \nonumber\\
    \hat{\Sigma} & \xleftarrow{} & \frac{(1 - \gamma) \Tilde{T} \hat{\Sigma} + Y_{T+1} \otimes Y_{T}}{(1 - \gamma) \Tilde{T} + 1 } ; \nonumber\\
    \hat{K} & \xleftarrow{} & \frac{(1 - \gamma) \Tilde{T} \hat{K} + Y_{T+1} \otimes Y_{T-1} \otimes Y_{T}}{(1 - \gamma) \Tilde{T} + 1 }  ; \nonumber\\
    \Tilde{T} & \xleftarrow{} & \Tilde{T} \cdot (1 - \gamma) + 1 . 
\end{eqnarray}
Here $\Tilde{T} = \sum_{i=1}^T (1 - \gamma)^{i - 1}$ serves as an effective sample size. This strategy is equivalent to calculating the exponentially weighted moving average for each parameter:
\begin{equation}
    \hat{\mu} = \frac{ \sum_{t=1}^T (1 - \gamma)^{T-t} Y_{t}  }{\sum_{t=1}^T (1 - \gamma)^{T-t} }  ; 
    \hat{\Sigma} = \frac{ \sum_{t=2}^{T} (1 - \gamma)^{T-t} Y_{t} \otimes Y_{t-1} }{\sum_{t=2}^{T} (1 - \gamma)^{T-t}} ; 
    \hat{K} = \frac{ \sum_{t=3}^{T} (1 - \gamma)^{T-t} Y_{t} \otimes Y_{t-2} \otimes Y_{t-1}  }{\sum_{t=3}^{T} (1 - \gamma)^{T-t}}  .
\end{equation}

\section{Simulations} \label{sec:simulations}

This section presents simulation studies to evaluate the performance of SHMM and PSHMM under various conditions. We assess robustness to different signal-to-noise ratios, misspecified models, and heavy-tailed data. We also test the effectiveness of the forgetfulness mechanism and compare computational efficiency.

\subsection{Tests of Robustness under Different Signal-Noise Ratio, Mis-specified Models and Heavy-Tailed Data}

\subsubsection{Experiment setting}
\label{subsubsec:experiment_setting}

We simulated $100$-dimensional data with $10000$ training points and $100$ test points using various GHMM configurations. Each simulation was repeated $100$ times. We compared SHMM, three variants of PSHMM (projection onto polyhedron, projection onto simplex, and online learning with projection onto simplex), and the E-M algorithm. Prediction performance was evaluated using the average $R^2$ over all repetitions.  The online learning variant of PSHMM used $1000$ training samples for the initial estimate (warm-up), and incorporated the remaining $9000$ training samples using online updates.  We also calculated an oracle $R^2$, assuming knowledge of all HMM parameters but not the specific hidden states.

In our simulations, online training for PSHMM differs from offline training for two reasons. First, the estimation of $\hat{U}$ and $\hat{M}$ are based only on the warm-up set for online learning (as is the case for the online version of SHMM), and the entire training set for offline learning. Second, during the training period, for PSHMM, the updated moments are based on the recursive predictions of $\hat{w}_t$, which are themselves based on the weights $ \{ w_s \}_{s=1}^{t-1}$.  In contrast, for the offline learning of PSHMM, the weights used to calculate the moments are based on the GMM estimates from the entire training dataset.  

For each state, we made two assumptions about the emission distribution. First, it has a one-hot mean vector in which the $i$-th state is $[\mathrm{1} \{ i = j \} ]_{j=1}^p$, where $\mathrm{1} \{ \cdot \}$ is the indicator function. Second, it has a diagonal covariance matrix.  We tested the SHMM, PSHMM methods, and the E-M algorithm under two types of transition matrix, five different signal-noise ratios and different emission distributions as below:
\begin{itemize}
    \item Transition matrix:
    \begin{itemize}
        \item Sticky: diagonal elements are $0.6$, off-diagonal elements are $\frac{0.4}{S-1}$, where $S$ is the number of states that generated the data;
        \item Non-sticky: diagonal elements are $0.4$, off-diagonal elements are $\frac{0.6}{S-1}$.
    \end{itemize}
    \item Signal-noise ratio: Covariance specified by $\sigma^2 I_{p}$, where $\sigma = \{0.01, 0.05, 0.1, 0.5, 1.0\}$.
     \item Emission distributions:
     \begin{itemize}
         \item Gaussian distribution, generated according to mean vector and covariance matrix;
         \item $t$ distribution: generated a standard a random vector of $i.i.d.$ $t_{5}$, $t_{10}$, $t_{15}$ and $t_{20}$ distribution, then multiplied by the covariance matrix and shifted by the mean vector. 
     \end{itemize}
\end{itemize} 

\subsubsection{Simulation Results}

\begin{figure}
    \centering
	\begin{subfigure}[t]{1\linewidth}
	    \centering
		\includegraphics[width=7in]{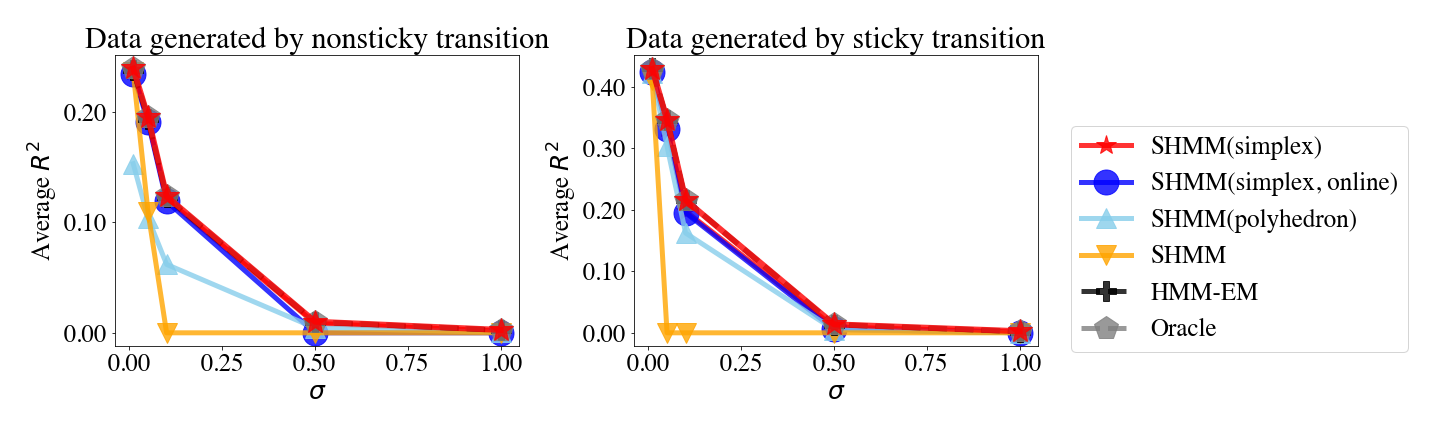}
  
         \vspace{-7mm}
	\end{subfigure}
    
     \begin{subfigure}[t]{1\linewidth}
	    \centering
		\includegraphics[width=7in]{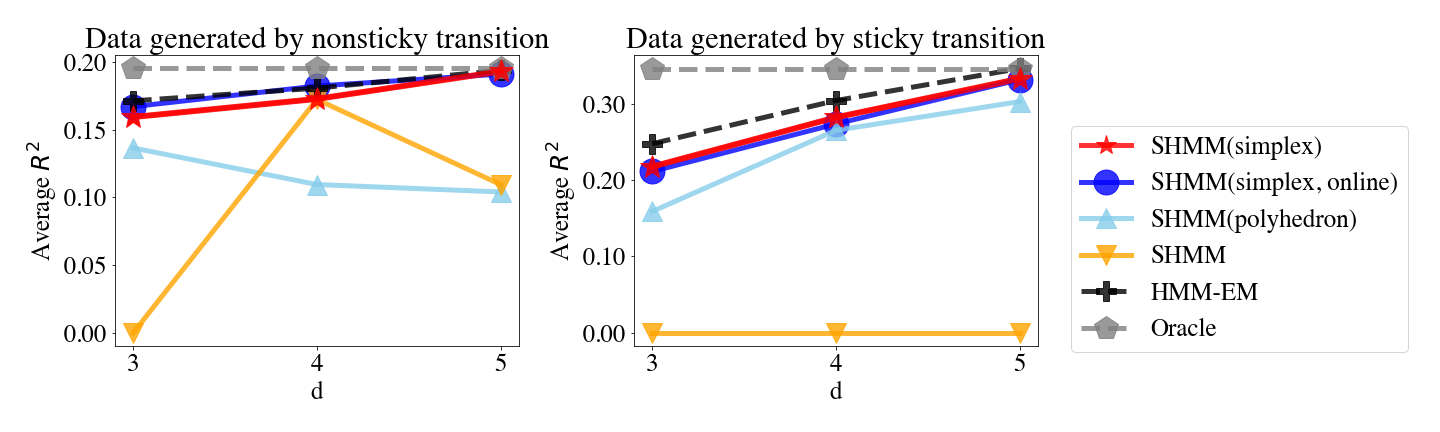}
  
         \vspace{-7mm}
	\end{subfigure}
 
     \begin{subfigure}[t]{1\linewidth}
	    \centering
		\includegraphics[width=7in]{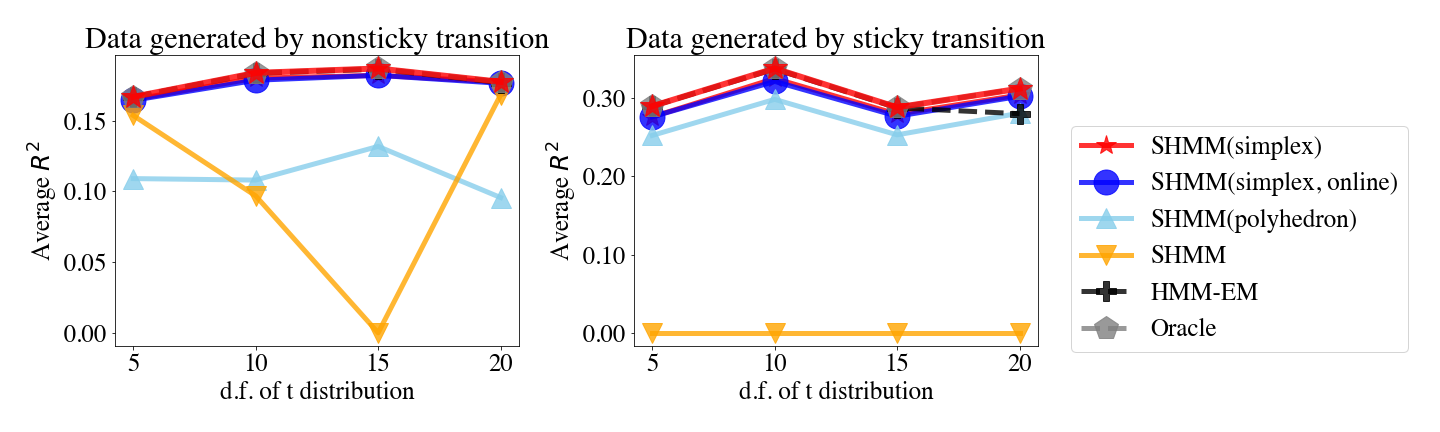}
  
         \vspace{-7mm}
	\end{subfigure}
 \caption{Comparative performance of SHMM, PSHMM, and E-M algorithms under various experimental conditions.
Left column: Results for sticky transitions. Right column: Results for non-sticky transitions.
Row 1: Effect of noise level $\sigma$ on algorithm performance (5-state GHMM).
Row 2: Impact of model misspecification - varying dimensionality d (5-state GHMM, $\sigma = 0.05$).
Row 3: Performance under non-Gaussian emissions (5-state HMM with t-distributed emissions, $\sigma = 0.05$, varying degrees of freedom).
Y-axis: Average $R^2$ (values $< 0$ thresholded to $0$ for visualization).
}
	\label{fig:simulation_results_6fig}
\end{figure}

Figure \ref{fig:simulation_results_6fig} shows that the performance of PSHMM is competitive with the E-M algorithm. While both methods achieve similar $R^2$ values in most scenarios, PSHMM outperforms E-M under conditions of high noise or heavy-tailed distributions. Note that our E-M implementation employs multiple initializations to mitigate local optima issues, which comes at the cost of increased computational time.

Our simulations used $100$-dimensional observed data ($dim = 100$), which we found approached the practical limits of the E-M algorithm. In higher-dimensional scenarios ($dim = 1000$ or $10000$), E-M failed, whereas SHMM continued to perform effectively. To ensure a fair comparison, we present results for $dim = 100$ throughout this section. 


The last point is important.  In Figure \ref{fig:simulation_results_6fig}, in order to include the E-M algorithm in our simulations, we restrict our investigation to settings where E-M is designed to succeed, and show comparable performance with our methodology.  Many real-world scenarios exist in a space where E-M is a nonstarter.  This highlights PSHMM's robustness and scalability advantages over traditional E-M approaches, particularly for high-dimensional data.

In Figure \ref{fig:simulation_results_6fig}, we can also see that adding projections to SHMM greatly improved $R^2$, achieving near oracle-level performance in some settings. The top row shows that, under both high- and low-signal-to-noise scenarios, PSHMM works well. The middle row shows that PSHMM is robust and outperforms SHMM when the model is mis-specified, for example, when the underlying data contains 5 states but we choose to reduce the dimensions to $3$ or $4$. The last row shows that PSHMM is more robust and has a better $R^2$ than SHMM with heavy-tailed data such as data generated by a t distribution. In all figures, when $R^2 < 0$, we threshold to $0$ for plotting purpose. See supplementary tables for detailed results. Negative $R^2$ occurs only for the standard SHMM, and implies that it is not stable.  Since we simulated $100$ trials for each setting, in addition to computing the mean $R^2$ we also calculated the variance of the $R^2$ metric.  We provide the variance of the $R^2$ in the appendix, but note here that the variance of $R^2$ under PSHMM was smaller than that of the SHMM.  Overall, while SHMM often performs well, it is not robust, and PSHMM provides a suitable solution. 

Finally, in all simulation settings, we see that the standard SHMM tends to give poor predictions except in non-sticky and high signal-to-noise ratio settings. PSHMM is more robust against noise and mis-specified models. Among the PSHMM variants, projection-onto-simplex outperforms projection-onto-polyhedron. The reason is that the projection-onto-simplex has a dedicated optimization algorithm that guarantees the optimal solution in the projection step. In contrast, projection-onto-polyhedron uses the log-barrier method, which is a general purpose optimization algorithm and does not guarantee the optimality of the solution. Since projection-onto-polyhedron also has a higher computational time, we recommend using projection-onto-simplex. Additionally, for projection-onto-simplex, we see that online and offline estimation perform similarly in most settings, suggesting that online learning does not lose too much power compared with offline learning. 

\subsection{Testing the effectiveness of forgetfulness}

\paragraph{Experimental setting.} Similarly to Section \ref{subsubsec:experiment_setting}, we simulated 100-dimensional, 5-state GHMM data with $\sigma = \{0.01, 0.05, 0.1, 0.5, 1.0\}$ and of length $2000$ where the first $1000$ steps are for training and the last $1000$ steps are for testing. The transition matrix is no longer time-constant but differ in the training and testing period as follows:
\begin{itemize}
    \item training period (diagonal-0.8): $\bm{T}^{(train)} = [\bm{T}_{ij}^{(train)}]_{i=1,\cdots,5}^{j=1,\cdots,5} = [0.75 \times \mathrm{I}\{ i= j \} + 0.05]_{i=1,\cdots,5}^{j=1,\cdots,5}$;
    \item testing period (antidiagonal-0.8): $\bm{T}^{(test)} = [\bm{T}_{ij}^{(test)}]_{i=1,\cdots,5}^{j=1,\cdots,5} = [0.75 \times \mathrm{I}\{ i+ j = 5 \} + 0.05]_{i=1,\cdots,5}^{j=1,\cdots,5}$.
\end{itemize}

We tested different methods on the last 100 time steps, including the standard SHMM, projection-onto-simplex SHMM,  online learning projection-onto-simplex SHMM, online learning projection-onto-simplex SHMM with decay factor $\gamma = 0.05$, E-M, and the oracle as defined above. For online learning methods, we used the first $100$ steps in the training set for warm-up and incorporated the remaining $900$ samples using online updates. 

\begin{figure}[!htb]
    \centering
    \includegraphics[width=6in]{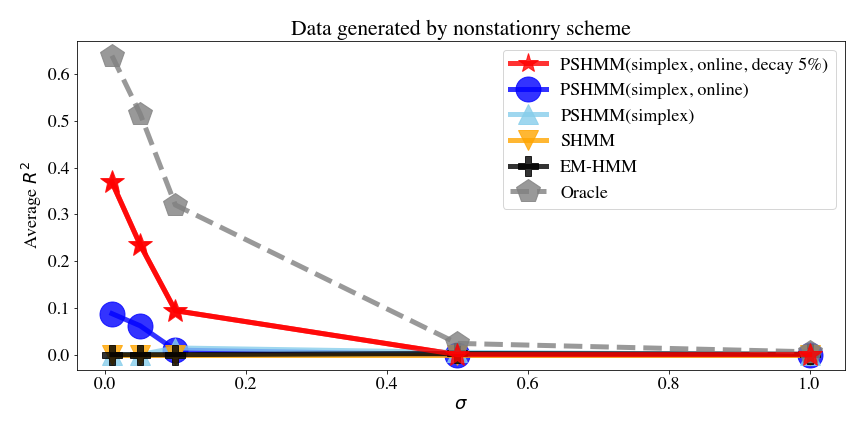}
 \caption{Simulation results for online learning variants. The simulations are generated using a 5-state GHMM across a variety of $\sigma$ settings and time-varying transition distributions. The y-axis is the average $R^2$. As before, for $R^2 < 0$, we threshold at $0$ for plotting purposes. See supplementary for detailed results.}
	\label{fig:simulation_decay}
\end{figure}

\paragraph{Simulation results.} Figure \ref{fig:simulation_decay} shows the simulation results. Since the underlying data generation process is no longer stationary, most methods failed including the E-M algorithm, with the exception of the online learning projection-onto-simplex SHMM with decay factor $= 0.05$. Adding the decay factor was critical for accommodating non-stationarity.  As $\sigma$ increases even PSHMM performs poorly-- the oracle shows that this degredation in performance is hard to overcome.

\subsection{Testing Computational Time}

\paragraph{Experimental setting} 
We used a similar experimental setting from Section \ref{subsubsec:experiment_setting}. We simulated 100-dimensional, 3-state GHMM data with $\sigma=0.05$ and length 2000. We used the first half for warm-up and tested computational time on the last 1000 time steps. We tested both E-M algorithm and SHMM (all variants). For the SHMM family of algorithms, we tested under both online and offline learning regimes. We computed the total running time in seconds. The implementation is done in python with packages Numpy \citep{harris2020numpy}, Scipy \citep{2020SciPy} and scikit-learn \citep{scikit-learn} without multithreading. Note that in contrast to our previous simulations, the entire process is repeated only 30 times. The computational time is the average over the 30 runs. 

\paragraph{Simulation results}
Table \ref{tab:simulation_time} shows the computational times for each method. First, online learning substantially reduced the computational cost. For the offline learning methods, projection-onto-simplex SHMM performed similarly to SHMM, and projection-onto-polyhedron was much slower.  In fact, the offline version of projection-onto-polyhedron was slow even compared to the the E-M algorithm. However, the online learning variant of projection-onto-polyhedron was much faster than the Baum-Welch algorithm. Taking both the computational time and prediction accuracy into consideration, we conclude that online and offline projection-onto-simplex SHMM are the best choices among these methods.  

\begin{table}[!htb]
    \centering
    \begin{tabular}{c|c|c}
    \hline
    Method & Offline/online & Computational time (sec)\\
    \hline
        E-M (Baum-Welch) & -	& 2134 \\
        SHMM	& offline &	304 \\
        SHMM	& online	& 0.5 \\
       PSHMM (simplex)	& offline	& 521 \\
       PSHMM (simplex)	& online	& 0.7 \\
       PSHMM (polyhedron)	& offline	& 10178 \\
       PSHMM (polyhedron)	& online	& 14 \\
        \hline
    \end{tabular}
    \caption{Simulation results for comparing computational time among different methods.}
    \label{tab:simulation_time}
\end{table}

\section{Application: Backtesting on High Frequency Crypto-Currency Data} \label{sec: application}

\subsection{Data Description \& Experiment Setting}

To evaluate our algorithm's performance on real-world data, we utilized a cryptocurrency trading dataset from Binance (https://data.binance.vision), one of the world's largest Bitcoin exchanges. We focused on minute-level data for five major cryptocurrencies: Bitcoin, Ethereum, XRP, Cardano, and Polygon. Our analysis used log returns for each minute as input for our models.

We used the period from July 1, 2022, to December 31, 2022, as our test set. For each day in this test set, we employed a 30-day rolling window of historical data for model training.  After training, we generated consecutive-minute recursive predictions for the entire test day without updating the model parameters intra-day.

Our study compared three models: the HMM with EM (HMM-EM), the SHMM, and the PSHMM with simplex projection. For all models, we assumed 4 latent states, motivated by the typical patterns observed in log returns, which often fall into combinations of large/small gains/losses.

To assess the practical implications of our models' predictions, we implemented a straightforward trading strategy. This strategy generated buy signals for positive forecasted returns and short-sell signals for negative forecasted returns. We simulated trading a fixed dollar amount for each cryptocurrency, holding positions for one minute, and executing trades at every minute of the day.

We evaluated the performance of our trading strategy using several standard financial metrics. The daily return was calculated as the average return across all five cryptocurrencies, weighted by our model's predictions:
$$ R_m = \frac{1}{5} \sum_{i=1}^5 \sum_{t} sign(\widehat{Y}_{i, t}^{(m)}) Y_{i, t}^{(m)} $$
where $Y_{i, t}^{(m)}$ is the return for minute $t$ of day $m$ for currency $i$, $\widehat{Y}_{i, t}^{(m)}$ is its prediction, and $sign(a)$ is $1$ if $a$ is positive, $-1$ if $a$ is negative, and $0$ if $a = 0$.  From these daily returns, we computed the annualized return
$$ Annualized \ return = 365 \times \overline{R}  , $$
to measure overall profitability. To assess risk-adjusted performance, we calculated the Sharpe ratio \citep{sharpe1966sharpe_ratio}
$$ Sharpe \ ratio   = \frac{ \sqrt{365} \times \overline{R}}{\widehat{std} (R)} , $$
where $\overline{R}$ and $\widehat{std} (R)$ are the sample mean and standard deviation of the daily returns.  The Sharpe ratio provides insight into the return earned per unit of risk. Lastly, we determined the maximum drawdown \citep{grossman1993mdd_1}
$$ Maximum \ drawdown = \max_{m_2  } \max_{m_1 < m_2} \left[ \frac{ \sum_{m = m_1}^{m_2} (- R_m) } {1 + \sum_{m = 1}^{m_1} R_m} \right] , $$ which quantifies the largest peak-to-trough decline as a percentage. Since the financial data is leptokurtic, the maximum drawdown shows the outlier effect better than the Sharpe ratio which is purely based on the first and second order moments. A smaller maximum drawdown indicates that the method is less risky.  

\subsection{Results}
\begin{table}[!htb]
    \centering
    \begin{tabular}{c|c|c|c}
        \hline
         Method  & Sharpe Ratio & Annualized Return & Maximum drawdown \\
         \hline
         PSHMM  & $2.88$ & $1012\%$ & $49\%$ \\
         SHMM  & $1.07$ & $345\%$ & $90\%$ \\
         HMM-EM  & $0.89$ & $197\%$ & $53\%$ \\
        \hline
    \end{tabular}
    \caption{Real-world application results: PSHMM, SHMM, HMM-EM and AR on crypto-currency trading.  }
    \label{tab:res_realdata_bitcoin}
\end{table}

From Table \ref{tab:res_realdata_bitcoin}, we see that PSHMM outperforms all other benchmarks with the highest Sharpe ratio and annualized return, and the lowest maximum drawdown. PSHMM outperforms SHMM and SHMM outperforms HMM-EM. SHMM outperforms HMM-EM because spectral learning does not suffer from the local minima problem of the E-M algorithm. PSHMM outperforms SHMM because projection-onto-simplex provides regularization. 

\begin{figure}[!htb]
    \centering
    \includegraphics[width=6in]{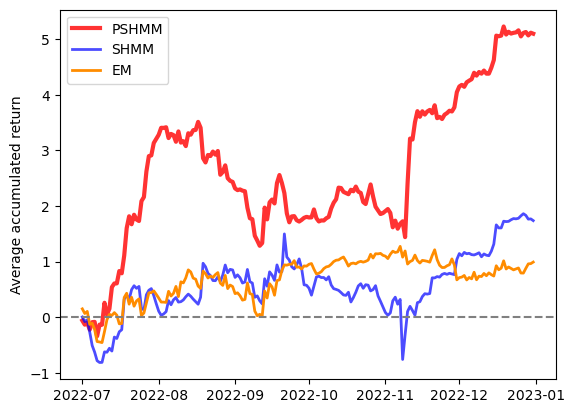}
    \caption{Comparative performance of PSHMM, SHMM, and HMM-EM in cryptocurrency trading.
The graph shows the average accumulated return over time for each method applied to five cryptocurrencies (Bitcoin, Ethereum, XRP, Cardano, and Polygon) from July to December 2022.
Y-axis: Average accumulated return. X-axis: Trading days.}
    \label{fig:res_realdata_bitcoin_cumreturn}
\end{figure}

The accumulated daily return is shown in Figure \ref{fig:res_realdata_bitcoin_cumreturn}. PSHMM outperformed the other methods.
The maximum drawdown of PSHMM is $49\%$. Considering the high volatility of the crypto-currency market during the second half of 2022, this maximum drawdown is acceptable. 
For computational purposes, the drawdown is allowed to be larger than $100\%$ because we always use a fixed amount of money to buy or sell, so we are effectively assuming an infinite pool of cash. Between PSHMM and SHMM, the only difference is the projection-onto-simplex. We see that the maximum drawdown of PSHMM is only about half that of SHMM, showing that PSHMM takes a relatively small risk, especially given that PSHMM has a much higher return than SHMM.

\section{Discussion} \label{sec:discussion}

\paragraph{Spectral estimation avoids being trapped in local optima.} E-M (i.e. the B-W algorithm) optimizes the likelihood function and is prone to local optima since the likelihood function is highly non-convex, whereas spectral estimation uses the MOM directly on the observations.  Although multiple initializations can mitigate the local optima problem with E-M, it, there is no guarantee that it will convergence to the true global optimum. Spectral estimation provides a framework for estimation which not only avoids non-convex optimization, but also has nice theoretical properties. The approximate error bound tells us that when the number of observations goes to infinity, the approximation error will go to zero. In this manuscript, we also provide the asymptotic distribution for this error.

\paragraph{Projection-onto-simplex serves as regularization.} The standard SHMM can give poor predictions due to the accumulation and propagation of errors. Projection-onto-simplex pulls the prediction back to a reasonable range. This regularization is our primary methodological innovation, and importantly makes the SHMM well-suited for practical use. Although the simplex, estimated by the means from a GMM, can be biased, this simplex provides the most natural and reasonable choice for a projection space. This regularization introduces a bias-variance trade-off. When the data size is small, it accepts some bias in order to reduce variance. 

\paragraph{Online learning can adapt to dynamic patterns and provide faster learning.} Finally, we provide an online learning strategy that allows the estimated moments to adapt over time, which is critical in several applications that can exhibit nonstationarity. Our online learning framework can be applied to both the standard SHMM and PSHMM. Importantly, online learning substantially reduces the computational costs compared to re-training the entire model prior to each new prediction.

\bigskip
\newpage
\appendix
\begin{center}
{\large\bf SUPPLEMENTAL MATERIALS}
\end{center}



\section{Proof for Lemma 1}
\begin{proof*}[Lemma 1]
First, we expand the estimated likelihood by decomposing it into the underlying truth plus the error terms. We have
\begin{eqnarray}
      & & \widehat Pr(x_{1:T})  \nonumber \\
      & = & \hat c_{\infty}^\top \hat C(y_T) \hat C(y_{T-1}) \cdots \hat C(y_1) \hat c_1 \nonumber \\
      & = &  (\hat \mu^\top \hat \Sigma^{-1}) [\hat K(y_T) \hat \Sigma^{-1}] [\hat K(y_{T-1}) \hat \Sigma^{-1}] \cdots [\hat K(y_1) \hat \Sigma^{-1}] \hat \mu \nonumber \\
      &  = & [(\mu + \widehat{\Delta \mu})^\top (\Sigma + \widehat{\Delta \Sigma})^{-1}] [(K + \widehat{\Delta K})(y_T) (\Sigma + \widehat{\Delta \Sigma})^{-1}] \nonumber \\
      & & \cdots [(K + \widehat{\Delta K})(y_1) (\Sigma + \widehat{\Delta \Sigma})^{-1}](\mu + \widehat{\Delta \mu}) . \nonumber \\
      \label{eqn:ErrorTermExpansion}
\end{eqnarray}

Consider the matrix perturbation $ (\Sigma + \widehat{\Delta \Sigma})^{-1} = \Sigma^{-1} - \Sigma^{-1} \widehat{\Delta \Sigma} \Sigma^{-1} + O( || \widehat{\Delta \Sigma} ||^2 )$, Here the matrix norm $||\cdot||$ can be any norm since all matrix norms have equivalent orders. Also note that all items with $\widehat{\Delta}$ are $O_p(N^{-\frac{1}{2}})$. $N$ is the number of $i.i.d.$ triple $(Y_1, Y_2, Y_3)$ for estimating $\widehat{\mu}, \widehat{\Sigma}, \widehat{K}$. Note that $N$ and $T$ are not related, and that we work in the regime where $T$ is fixed but $N \xrightarrow{} \infty$. For example, $\hat{\mu} = \mu + \widehat{\Delta \mu} = \frac{1}{N} \Sigma_{i=1}^N Y_i \sim MVN(\mu, \frac{1}{N} Cov(Y)) $. Similar analyses apply to $\widehat{\Sigma}$ and $\widehat{K}$ can be similarly. So 
\begin{eqnarray*}
    \widehat{\Delta \mu} & = & [O_p (N^{-1/2})]^{(d)} ;\\
    \widehat{\Delta \Sigma} & = & [O_p (N^{-1/2})]^{(d \times d)} ; \\ 
    \widehat{\Delta K} & = & [O_p (N^{-1/2})]^{(d \times d \times d)} ,
\end{eqnarray*}
where $d$ is the dimension of $Y_t$. One application is $ (\Sigma + \widehat{\Delta \Sigma})^{-1} = \Sigma^{-1} - \Sigma^{-1} \widehat{\Delta \Sigma} \Sigma^{-1} + O_p ( N^{-1} )$.

According to the previous two expansions, we can rewrite Eq \ref{eqn:ErrorTermExpansion} as
\begin{eqnarray*}
      & &  [(\mu + \widehat{\Delta \mu})^\top (\Sigma + \widehat{\Delta \Sigma})^{-1}] [(K + \widehat{\Delta K})(y_T) (\Sigma + \widehat{\Delta \Sigma})^{-1}] \cdots \\
      & & [(K + \widehat{\Delta K})(y_1) (\Sigma + \widehat{\Delta \Sigma})^{-1}](\mu + \widehat{\Delta \mu})   \\
      &  = & [(\mu + \widehat{\Delta \mu})^T (\Sigma^{-1} - \Sigma^{-1} \widehat{\Delta \Sigma} \Sigma^{-1} + O_p( N^{-1} ))] \\
      & & [(K + \widehat{\Delta K})(y_T) (\Sigma^{-1} - \Sigma^{-1} \widehat{\Delta \Sigma} \Sigma^{-1} + O_p( N^{-1} ))] \cdots \\
      & & [(K + \widehat{\Delta K})(y_1) (\Sigma^{-1} - \Sigma^{-1} \widehat{\Delta \Sigma} \Sigma^{-1} + O_p( N^{-1} ))](\mu + \widehat{\Delta \mu}) \\
      & = & \mu^\top \Sigma^{-1} K(y_T) \Sigma^{-1} \cdots K(y_1) \Sigma^{-1} \mu  + (v+ \tilde{v})^\top \widehat{\Delta \mu} + \sum_{t=1}^T a_t^\top \widehat{\Delta K}(y_t) \tilde{a_t} \\
      & & - \sum_{t=0}^T b_t^\top \widehat{\Delta \Sigma} \tilde{b_t} + O_p (N^{-1}) \\
      & = & Pr(x_{1:T})   + (v+ \tilde{v})^\top \widehat{\Delta \mu} + \sum_{t=1}^T a_t^\top \widehat{\Delta K}(y_t) \tilde{a_t} - \sum_{t=0}^T b_t^\top \widehat{\Delta \Sigma} \tilde{b_t} + O_p (N^{-1})
\end{eqnarray*}
In the above, we first substitute $(\Sigma + \widehat{\Delta \Sigma})^{-1}$ with $\Sigma^{-1} - \Sigma^{-1} \widehat{\Delta \Sigma} \Sigma^{-1} + O_p ( N^{-1} )$, then distribute all multiplications. After distribution, we can categorize the resulting terms into 3 categories according to different orders of convergence. The first category is the deterministic term without any randomness or estimation error. There is only one term in this category, which is $Pr(x_{1:T})$, i.e. $\mu^\top \Sigma^{-1} K(y_T) \Sigma^{-1} \cdots K(y_1) \Sigma^{-1} \mu$. The second category is the part that converges with order $O_p( N^{-1/2} )$. This category involves all terms with one `$\widehat\Delta$' term, i.e. one $\widehat{\Delta \mu}$, $ \widehat{\Delta \Sigma} $ or $\widehat{\Delta K}$. These terms comprise the sum $ (v+ \tilde{v})^\top \widehat{\Delta \mu} + \sum_{t=1}^T a_t^\top \widehat{\Delta K}(y_t) \tilde{a_t} - \sum_{t=0}^T b_t^\top \widehat{\Delta \Sigma} \tilde{b_t}$. To be useful for the main Theorem, we simplify each term by representing the collection of ``true'' quantities that exist in each term with $v, \tilde{v}, a_t, \tilde{a}_t, b_t, \tilde{b}_t$ as defined in the lemma. Note that each collection contains only terms that fall either to the left or to the right of the `$\widehat\Delta$' term. 
 The third category contains all remaining terms. These terms converge faster than-- or on the order of-- $O_p ( N^{-1} )$. These terms are all finite, and thus their summation is also $O_p ( N^{-1} )$. 


\qed
\end{proof*}

\section{Detailed Results for Simulations}
Table \ref{tab:simulation_details} shows the detailed simulation results. The first columns represents the emission distribution. The second column shows the type of the transition matrix, which we describe in the simulation settings of the paper. Here we show two types of oracle: ``limited oracle'' is the oracle described in the paper (and is the most commonly used oracle). Limited oracle assumes we know all parameters but don't know the true hidden states. The ``strong oracle'' assumes knowledge of every parameter, as well as the particular hidden state at each time point. From the table below, we can see that these oracles nearly overlapped, and that PSHMM with projection-onto-simplex is very close to the oracle in most cases.

\begin{sidewaystable}[!htp]
    \centering
    \begin{tabular}{c|c|c|c|c|c|c|c|c|c}
    \hline
   E & T & $\sigma$ & \#Cluster & Limited & Strong & SHMM & PSHMM & PSHMM  & PSHMM \\

    & & & inferred & oracle & oracle &  & simplex & polyhedron & simplex\\
    & & & & & & & & & online \\
    \hline
$t_5$ & sticky & $0.05$ & $5$ & $0.28$ & $0.28$ & $-673629.23$ & $0.27$ & $0.24$ & $0.26$ \\
$t_{10}$ & sticky & $0.05$ & $5$ & $0.3$ & $0.3$ & $-67.72$ & $0.29$ & $0.27$ & $0.29$ \\
$t_{15}$ & sticky & $0.05$ & $5$ & $0.31$ & $0.31$ & $-2.82$ & $0.3$ & $0.28$ & $0.29$ \\
$t_{20}$ & sticky & $0.05$  & $5$ & $0.3$ & $0.3$ & $-205.24$ & $0.29$ & $0.27$ & $0.29$ \\
$t_{5}$ & nonsticky & $0.05$ & $5$ & $0.17$ & $0.17$ & $-2.48$ & $0.17$ & $0.11$ & $0.17$ \\
$t_{10}$ & nonsticky & $0.05$  & $5$ & $0.19$ & $0.19$ & $0.11$ & $0.18$ & $0.11$ & $0.18$ \\
$t_{15}$ & nonsticky & $0.05$  & $5$ & $0.19$ & $0.19$ & $-0.59$ & $0.18$ & $0.11$ & $0.18$  \\
$t_{20}$ & nonsticky & $0.05$  & $5$ & $0.19$ & $0.19$ & $0.14$ & $0.19$ & $0.11$ & $0.18$ \\
$N$ & sticky & $0.05$ & $3$ & $0.31$ & $0.31$ & $-125.03$ & $0.21$ & $0.12$ & $0.21$ \\
$N$ & sticky & $0.05$ & $4$ & $0.31$ & $0.31$ & $-717.8$ & $0.25$ & $0.23$ & $0.25$ \\
$N$ & nonsticky & $0.05$ & $3$ & $0.19$ & $0.19$ & $-3.25$ & $0.16$ & $0.1$ & $0.17$ \\
$N$ & nonsticky & $0.05$  & $4$ & $0.19$ & $0.19$ & $-155.85$ & $0.17$ & $0.09$ & $0.18$ \\
$N$ & sticky & $0.01$  & $5$ & $0.38$ & $0.38$ & $0.34$ & $0.38$ & $0.37$ & $0.37$ \\
$N$ & nonsticky & $0.01$  & $5$ & $0.24$ & $0.24$ & $0.24$ & $0.24$ & $0.15$ & $0.24$ \\
$N$& sticky & $0.05$ & $5$ & $0.31$ & $0.31$ & $-9.36$ & $0.3$ & $0.28$ & $0.3$ \\
$N$ & nonsticky & $0.05$ & $5$ & $0.19$ & $0.19$ & $0.14$ & $0.19$ & $0.12$ & $0.19$  \\
$N$ & sticky & $0.1$ & $5$ & $0.19$ & $0.19$ & $-74.73$ & $0.18$ & $0.15$ & $0.18$ \\
$N$ & sticky & $0.5$  & $5$ & $0.01$ & $0.01$ & $-22.02$ & $0.01$ & $0.0$ & $0.0$ \\
$N$ & sticky & $1.0$  & $5$ & $0.0$ & $0.0$ & $-1163.8$ & $0.0$ & $-0.0$ & $-0.01$ \\
$N$ & nonsticky & $0.1$  & $5$ & $0.12$ & $0.12$ & $-8.47$ & $0.12$ & $0.07$ & $-0.0$ \\
$N$ & nonsticky & $0.5$ & $5$ & $0.01$ & $0.01$ & $-22.29$ & $0.01$ & $0.0$ & $-0.0$ \\
$N$ & nonsticky & $1.0$  & $5$ & $0.0$ & $0.0$ & $-5.61$ & $0.0$ & $-0.0$ & $-0.01$ \\
        \hline
    \end{tabular}
    \caption{Detailed simulation results.}
    \label{tab:simulation_details}
\end{sidewaystable}

\bibliographystyle{apalike}
\bibliography{reference}
\end{document}